\documentclass{article}

\usepackage{microtype}
\usepackage{graphicx}
\usepackage{booktabs} 

\usepackage{hyperref}
\usepackage{latexsym}
\usepackage{multirow}
\usepackage{xspace}
\usepackage{caption}
\usepackage{subcaption}

\newcommand{\task}{\textsc{xtreme}\xspace}

\usepackage[accepted]{icml2020}

\icmltitlerunning{XTREME: A Benchmark for Evaluating Cross-lingual generalization}

\begin{document}

\twocolumn[
\icmltitle{XTREME: A Massively Multilingual Multi-task Benchmark\\for Evaluating Cross-lingual Generalization}



\icmlsetsymbol{equal}{*}

\begin{icmlauthorlist}
\icmlauthor{Junjie Hu}{equal,cmu}
\icmlauthor{Sebastian Ruder}{equal,dm}
\icmlauthor{Aditya Siddhant}{goo}
\icmlauthor{Graham Neubig}{cmu}
\icmlauthor{Orhan Firat}{goo}
\icmlauthor{Melvin Johnson}{goo}
\end{icmlauthorlist}

\icmlaffiliation{cmu}{Carnegie Mellon University}
\icmlaffiliation{dm}{DeepMind}
\icmlaffiliation{goo}{Google Research}

\icmlcorrespondingauthor{Junjie Hu}{\texttt{junjieh@cs.cmu.edu}}
\icmlcorrespondingauthor{Melvin Johnson}{\texttt{melvinp@google.com}}

\icmlkeywords{Machine Learning, ICML}

\vskip 0.3in
]



\printAffiliationsAndNotice{\icmlEqualContribution} 

\begin{abstract}
Much recent progress in applications of machine learning models to NLP has been driven by benchmarks that evaluate models across a wide variety of tasks. However, these broad-coverage benchmarks have been mostly limited to English, and despite an increasing interest in multilingual models, a benchmark that enables the comprehensive evaluation of such methods on a diverse range of languages and tasks is still missing.
To this end, we introduce the Cross-lingual TRansfer Evaluation of Multilingual Encoders (\task) benchmark, a multi-task benchmark for evaluating the cross-lingual generalization capabilities of multilingual representations across 40 languages and 9 tasks. We demonstrate that while models tested on English reach human performance on many tasks, there is still a sizable gap in the performance of cross-lingually transferred models, particularly on syntactic and sentence retrieval tasks. There is also a wide spread of results across languages. We release the benchmark\footnote{The benchmark is publicly available at \url{https://sites.research.google/xtreme}. The codes used for downloading data and training baseline models are available at \url{https://github.com/google-research/xtreme}.} to encourage research on cross-lingual learning methods that transfer linguistic knowledge across a diverse and representative set of languages and tasks.
\end{abstract}

\section{Introduction}

In natural language processing (NLP), there is a pressing urgency to build systems that serve \emph{all} of the world's approximately 6,900 languages to overcome language barriers and enable universal information access for the world's citizens \cite{Ruder2019survey,aharoni2019massively, Arivazhagan2019massively_multilingual}.
At the same time, building NLP systems for most of these languages is challenging due to a stark lack of data.
Luckily, many languages have similarities in syntax or vocabulary, and multilingual learning approaches that train on multiple languages while leveraging the shared structure of the input space have begun to show promise as ways to alleviate data sparsity. Early work in this direction focused on single tasks, such as grammar induction \cite{Snyder2009unsupervised}, part-of-speech (POS) tagging \cite{Tackstrom2013token_and_type}, parsing \cite{McDonald2011delexicalized}, and text classification \cite{Klementiev2012inducing}. Over the last few years, there has been a move towards \emph{general-purpose multilingual representations} that are applicable to many tasks, both
on the word level \cite{mikolov2013exploiting,faruqui2014improving,artetxe2017learning} or the full-sentence level \cite{Devlin2019bert,Lample2019xlm}. Despite the fact that such representations are intended to be general-purpose, evaluation of them has often been performed on a very limited and often disparate set of tasks---typically focusing on translation \cite{Glavas2019,Lample2019xlm} and classification \cite{Schwenk2018mldoc,Conneau2018xnli}---and typologically similar languages \cite{conneau2018word}.

To address this problem and incentivize research on truly general-purpose cross-lingual representation and transfer learning, we introduce the Cross-lingual TRansfer Evaluation of Multilingual Encoders (\task) benchmark. \task covers 40 typologically diverse languages spanning 12 language families and includes 9 tasks that require reasoning about different levels of syntax or semantics.\footnote{By typologically diverse, we mean languages that span a wide set of linguistic phenomena such as compounding, inflection, derivation, etc.~which occur in many of the world's languages.} In addition, we introduce \emph{pseudo} test sets as diagnostics that cover all 40 languages by automatically translating the English test set of the natural language inference and question-answering dataset to the remaining languages. 

\task focuses on the \emph{zero-shot cross-lingual transfer} scenario, where annotated training data is provided in English but none is provided in the language to which systems must transfer.%
\footnote{This is done both for efficiency purposes (as it only requires testing, not training, on each language) and practical considerations (as annotated training data is not available for many languages).}
We evaluate a range of state-of-the-art machine translation (MT) and multilingual representation-based approaches to performing this transfer.
We find that while state-of-the-art models come close to human performance in English on many of the tasks we consider, performance drops significantly when evaluated on other languages.
Overall, performance differences are highest for syntactic and sentence retrieval tasks.
Further, while models do reasonably well in most languages in the Indo-European family, we observe lower performance particularly for Sino-Tibetan, Japonic, Koreanic, and Niger-Congo languages. 

In sum, our contributions are the following: (i) We release a suite of 9 cross-lingual benchmark tasks covering 40 typologically diverse languages. (ii) We provide an online platform and leaderboard for the evaluation of multilingual models. (iii) We provide a set of strong baselines, which we evaluate across all tasks, and release code to facilitate adoption. (iv) We provide an extensive analysis of limitations of state-of-the-art cross-lingual models.

\section{Related Work}

\noindent \textbf{Cross-lingual representations} $\:$ Early work focused on learning cross-lingual representations using either parallel corpora \cite{gouws2015bilbowa,luong2015bilingual} or a bilingual dictionary to learn a linear transformation \cite{mikolov2013exploiting,faruqui2014improving}. Later approaches reduced the amount of supervision required using self-training \cite{artetxe2017learning} and unsupervised strategies such as adversarial training \cite{conneau2018word}, heuristic initialisation \cite{artetxe2018robust}, and optimal transport \cite{zhang2017earth}. Building on advances in monolingual transfer learning \cite{mccann2017learned,howard2018universal,peters2018deep,Devlin2019bert}, multilingual extensions of pretrained encoders have recently been shown to be effective for learning deep cross-lingual representations \cite{eriguchi2018zero,Pires2019,Wu2019,Lample2019xlm,Siddhant2019evaluating}. 

\noindent \textbf{Cross-lingual evaluation} $\:$ One pillar of the evaluation of cross-lingual representations has been translation, either on the word level (\emph{bilingual lexicon induction}) or on the sentence level (\emph{machine translation}). In most cases, evaluation has been restricted to typologically related languages and similar domains; approaches have been shown to fail in less favorable conditions \cite{Glavas2019,Vulic2019,Guzman2019}. Past work has also reported issues with common datasets for bilingual lexicon induction \cite{czarnowska2019dont,kementchedjhieva2019lost} and a weak correlation with certain downstream tasks \cite{Glavas2019}. Translation, however, only covers one facet of a model's cross-lingual generalization ability. For instance, it does not capture differences in classification performance that are due to cultural differences \cite{mohammad2016translation,smith2016does}.

On the other hand, cross-lingual approaches have been evaluated on a wide range of tasks, including dependency parsing \cite{Schuster2019}, named entity recognition \cite{Rahimi2019}, sentiment analysis \citep{barnes2018bilingual}, natural language inference \cite{Conneau2018xnli}, document classification \cite{Schwenk2018mldoc}, and question answering \cite{artetxe2019cross,Lewis2019mlqa}. Evaluation on a single task is problematic as past work has noted potential issues with standard datasets: MLDoc \cite{Schwenk2018mldoc} can be solved by matching keywords \cite{artetxe2019cross}, while MultiNLI, the dataset from which XNLI \cite{Conneau2018xnli} was derived, contains superficial cues that can be exploited \cite{Gururangan2018}. Evaluation on multiple tasks is thus necessary to fairly compare cross-lingual models.  Benchmarks covering multiple tasks like GLUE \cite{Wang2019glue} and SuperGLUE \cite{wang2019superglue} have arguably spurred research in monolingual transfer learning. In the cross-lingual setting, such a benchmark not only needs to cover a diverse set of tasks but also languages. \task aims to fill this gap.

\section{\task}

\subsection{Design principles}
\label{sec:design}

\begin{table*}[]
\centering
\caption{Characteristics of the datasets in \task for the zero-shot transfer setting. For tasks that have training and dev sets in other languages, we only report the English numbers. We report the number of test examples per target language and the nature of the test sets (whether they are translations of English data or independently annotated). The number in brackets is the size of the intersection with our selected languages. For NER and POS, sizes are in sentences. Struct. pred.: structured prediction. Sent. retrieval: sentence retrieval.}
\resizebox{\textwidth}{!}{%
\begin{tabular}{l l r r r r r l l l}
\toprule
Task & Corpus & $|$Train$|$ & $|$Dev$|$ & $|$Test$|$ & Test sets & $|$Lang.$|$ & Task & Metric & Domain \\ \midrule
\multirow{2}{*}{Classification} & XNLI & 392,702 & 2,490 & 5,010 & translations & 15 & NLI & Acc. & Misc.  \\
& PAWS-X & 49,401 & 2,000 & 2,000 & translations & 7 & Paraphrase & Acc. & Wiki / Quora \\ \midrule
\multirow{2}{*}{Struct. pred.} & POS & 21,253 & 3,974 & 47-20,436 & ind. annot. & 33 (90) & POS & F1 & Misc. \\
& NER & 20,000 & 10,000 & 1,000-10,000 & ind. annot. & 40 (176) & NER & F1 & Wikipedia \\
\midrule
\multirow{3}{*}{QA} & XQuAD & \multirow{2}{*}{87,599} & \multirow{2}{*}{34,726} & 1,190 & translations & 11 & Span extraction & F1 / EM & Wikipedia \\
& MLQA &  &  & 4,517--11,590 & translations & 7 & Span extraction & F1 / EM & Wikipedia \\ 
& TyDiQA-GoldP & 3,696 & 634 & 323--2,719 & ind. annot. & 9 & Span extraction & F1 / EM & Wikipedia \\ \midrule
\multirow{2}{*}{Retrieval} & BUCC & - & - & 1,896--14,330 & - & 5 & Sent. retrieval & F1 & Wiki / news \\
& Tatoeba & - & - & 1,000 & - & 33 (122) & Sent. retrieval & Acc. & misc. \\
\bottomrule
\end{tabular}%
}
\label{tab:tasks}
\end{table*}

Given \task's goal of providing an accessible benchmark for the evaluation of cross-lingual transfer learning on a diverse and representative set of tasks and languages, we select the tasks and languages that make up the benchmark based on the following principles:

\noindent \textbf{Task difficulty} $\:$ Tasks should be sufficiently challenging so that cross-language performance falls short of human performance.

\noindent \textbf{Task diversity} $\:$ Tasks should require multilingual models to transfer their meaning representations at different levels, e.g.~words, phrases and sentences. For example, while classification tasks require sentence-level transfer of meaning, sequence labeling tasks like part-of-speech (POS) tagging or named entity recognition (NER) test the model's transfer capabilities at the word level.

\noindent \textbf{Training efficiency} $\:$ Tasks should be trainable on a single GPU for less than a day. This is to make the benchmark accessible, in particular to practitioners working with low-resource languages under resource constraints.

\noindent \textbf{Multilinguality} $\:$ We prefer tasks that cover as many languages and language families as possible.

\noindent \textbf{Sufficient monolingual data} $\:$ Languages should have sufficient monolingual data for learning useful pre-trained representations.

\noindent \textbf{Accessibility} $\:$ Each task should be available under a permissive license that allows the use and redistribution of the data for research purposes.

\subsection{Tasks}

\task consists of nine tasks that fall into four different categories requiring reasoning on different levels of meaning. 
We give an overview of all tasks in Table \ref{tab:tasks}, and describe the task details as follows.





\noindent \textbf{XNLI} $\:$ The Cross-lingual Natural Language Inference corpus \cite{Conneau2018xnli} asks whether a premise sentence entails, contradicts, or is neutral toward a hypothesis sentence. Crowd-sourced English data is translated to ten other languages by professional translators and used for evaluation, while the MultiNLI \cite{Williams2018multinli} training data is used for training.

\noindent \textbf{PAWS-X} $\:$ The Cross-lingual Paraphrase Adversaries from Word Scrambling \cite{Yang2019paws-x} dataset requires to determine whether two sentences are paraphrases. A subset of the PAWS dev and test sets \cite{Zhang2019paws} was translated to six other languages by professional translators and is used for evaluation, while the PAWS training set is used for training.


\noindent \textbf{POS} $\:$ We use POS tagging data from the Universal Dependencies v2.5 \cite{nivre2018universal} treebanks, which cover 90 languages. Each word is assigned one of 17 universal POS tags. We use the English training data for training and evaluate on the test sets of the target languages.


\noindent \textbf{NER} $\:$ For NER, we use the \texttt{Wikiann} \cite{Pan2017} dataset. Named entities in Wikipedia were automatically annotated with \texttt{LOC}, \texttt{PER}, and \texttt{ORG} tags in IOB2 format using a combination of knowledge base properties, cross-lingual and anchor links, self-training, and data selection. 
We use the balanced train, dev, and test splits from \citet{Rahimi2019}. 



\noindent \textbf{XQuAD} $\:$ The Cross-lingual Question Answering Dataset \cite{artetxe2019cross} requires identifying the answer to a question as a span in the corresponding paragraph. A subset of the English SQuAD v1.1 \cite{Rajpurkar2016squad} dev set was translated into ten other languages by professional translators and is used for evaluation.

\noindent \textbf{MLQA} $\:$ The Multilingual Question Answering \cite{Lewis2019mlqa} dataset is another cross-lingual question answering dataset similar to XQuAD. The evaluation data for English and six other languages was obtained by automatically mining target language sentences that are parallel to sentences in English from Wikipedia, crowd-sourcing annotations in English, and translating the question and aligning the answer spans in the target languages. For both XQuAD and MLQA, we use the SQuAD v1.1 training data for training and evaluate on the test data of the corresponding task.

\noindent \textbf{TyDiQA-GoldP} $\:$ We use the gold passage version of the Typologically Diverse Question Answering \cite{Clark2020tydiqa} dataset, a benchmark for information-seeking question answering, which covers nine languages. The gold passage version is a simplified version of the primary task, which uses only the gold passage as context and excludes unanswerable questions. It is thus similar to XQuAD and MLQA, while being more challenging as questions have been written without seeing the answers, leading to $3\times$ and $2\times$ less lexical overlap compared to XQuAD and MLQA respectively. We use the English training data for training and evaluate on the test sets of the target languages.


\noindent \textbf{BUCC} $\:$ The goal of the second and third shared task of the workshop on Building and Using Parallel Corpora  \cite{zweigenbaum2017overview,zweigenbaum2018overview} is to extract parallel sentences from a comparable corpus between English and four other languages. The dataset provides train and test splits for each language. For simplicity, we evaluate representations on the test sets directly without fine-tuning and calculate similarity using cosine similarity.\footnote{Results can be improved using more sophisticated similarity metrics \cite{Artetxe2019massively}.}


\noindent \textbf{Tatoeba} $\:$ We use the Tatoeba dataset \cite{Artetxe2019massively}, which consists of up to 1,000 English-aligned sentence pairs covering 122 languages. We find the nearest neighbour using cosine similarity and calculate error rate.

\subsection{Languages}

As noted in Section \ref{sec:design}, we choose our target languages based on availability of monolingual data, and typological diversity.
We use the number of articles in Wikipedia as a proxy for the amount of monolingual data available online. In order to strike a balance between language diversity and availability of monolingual data, we select all languages out of the top 100 Wikipedias\footnote{\url{https://meta.wikimedia.org/wiki/List_of_Wikipedias}} with the most articles as of December 2019.\footnote{This also has the benefit that they are covered by state-of-the-art methods such as mBERT and XLM.} We first select all languages that appear in at least three of our benchmark datasets. This leaves us with 19 languages, most of which are Indo-European or major world languages. We now select 21 additional languages that appear in at least one dataset and come from less represented language families. Wherever possible, we choose at least two languages per family.\footnote{For the Austro-Asiatic, Kartvelian, and Kra-Dai families as well as for isolates, we only obtain one language.}

In total, \task covers the following 40 languages (shown with their ISO 639-1 codes for brevity) belonging to 12 language families and two isolates: af, ar, bg, bn, de, el, en, es, et, eu, fa, fi, fr, he, hi, hu, id, it, ja, jv, ka, kk, ko, ml, mr, ms, my, nl, pt, ru, sw, ta, te, th, tl, tr, ur, vi, yo, and zh.
We provide a detailed overview of these languages in terms of their number of Wikipedia articles, linguistic features, and coverage in \task in the appendix.

While \task covers these languages in the sense that there is gold standard data in at least one task in each language, this does not mean that it covers all aspects of each language that are necessary for transfer. Languages may reveal different characteristics based on the task, domain, and register in which they are used. \task thus only serves as a glimpse into a model's true cross-lingual generalization capability.

\subsection{Pseudo test data for analyses}

\task covers 40 languages overall. Evaluation across the majority of languages is only possible for a subset of tasks, i.e. POS, NER, and Tatoeba.
As additional diagnositics and to enable a broader comparison across languages for a more diverse set of tasks, we automatically translate the English portions of a representative classification and QA task to the remaining languages using an in-house translation system.\footnote{Details of our translation system are provided in the appendix.} We choose XNLI and XQuAD as both have test sets that are translations of the English data by professional translators.

We first verify that performance on the translated test sets is a good proxy for performance on the gold standard test sets. We report the detailed results in the appendix. For XQuAD, the automatically translated test sets underestimate mBERT's true performance by 3.0 F1 / 0.2 EM points, similar to the 2.6 F1 points reported by \citet{Agic2018baselines} when translating the test data to other languages.\footnote{Note that even human translated test sets may underestimate a model's true cross-lingual generalization ability as such \emph{translationese} has been shown to be less lexically diverse than naturally composed language \cite{koppel2011translationese}.} For XNLI, the automatically translated test sets overestimate the true prediction accuracy by 2.4 points. In order to measure the translation quality between the human-translated test data and our pseudo test data, we compute the BLEU score, and the chrF score~\cite{popovic-2015-chrf}, which is suitable for measuring the translation quality of some languages such as Chinese and Russian. For the 14 languages in XNLI, we obtain average scores of 34.2 BLEU and 58.9 chrF scores on our pseudo test data compared to the reference translations, which correlate with a Pearson's $\rho$ of 0.57 and 0.28 respectively with mBERT performance.

Translating the English data to the remaining languages yields 40-way parallel pseudo test data that we employ for analyses in Section \ref{sec:analyses}.

\section{Experiments}

\subsection{Training and evaluation setup}

\task focuses on the evaluation of multilingual representations. We do not place any restriction on the amount or nature of the monolingual data used for pretraining multilingual representations. However, we request authors to be explicit about the data they use for training, in particular any cross-lingual signal. In addition, we suggest authors should not use any additional labelled data in the target task beyond the one that is provided.

For evaluation, we focus on \emph{zero-shot cross-lingual transfer} with English as the source language as this is the most common setting for the evaluation of multilingual representations and as many tasks only have training data available in English. Although English is not generally the best source language for cross-lingual transfer for all target languages \cite{Lin2019transfer_languages}, this is still the most practically useful setting. A single source language also facilitates evaluation as models only need to be trained once and can be evaluated on all other languages.\footnote{Future work may also consider multi-source transfer, which is interesting particularly for low-resource languages, and transfer to unknown languages or unknown language-task combinations.}

Concretely, pretrained multilingual representations are fine-tuned on English labelled data of an \task task. The model is then evaluated on the test data of the task in the target languages.

\subsection{Baselines}

We evaluate a number of strong baselines and state-of-the-art models. The approaches we consider learn multilingual representations via self-supervision or leverage translations---either for representation learning or for training models in the source or target language. We focus on models that learn deep contextual representations as these have achieved state-of-the-art results on many tasks. For comparability among the representation learning approaches, we focus on models that learn a multilingual embedding space between all languages in \task. We encourage future work to focus on these languages to capture as much language diversity as possible. We report hyper-parameters in the appendix. All hyper-parameter tuning is done on English validation data. We encourage authors evaluating on \task to do the same.

\noindent \textbf{mBERT} $\:$ Multilingual BERT \cite{Devlin2019bert} is a transformer model \cite{Vaswani2017attention} that has been pretrained on the Wikipedias of 104 languages using masked language modelling (MLM).

\noindent \textbf{XLM} $\:$ XLM \cite{Lample2019xlm} uses a similar pretraining objective as mBERT with a larger model, a larger shared vocabulary, and trained on the same Wikipedia data covering 100 languages.

\noindent \textbf{XLM-R} $\:$ XLM-R Large \cite{Conneau2019xlm-r} is similar to XLM but was trained on more than a magnitude more data from the web covering 100 languages.

\noindent \textbf{MMTE} $\:$ The massively multilingual translation encoder is part of an NMT model that has been trained on in-house parallel data of 103 languages extracted from the web \cite{Arivazhagan2019massively_multilingual}. For transfer, we fine-tune the encoder of the model \cite{Siddhant2019evaluating}.


\noindent \textbf{Translate-train} $\:$ For many language pairs, an MT model may be available, which can be used to obtain data in the target language. To evaluate the impact of using such data, we translate the English training data into the target language using our in-house MT system. We then fine-tune mBERT on the translated data. We provide details on how we align answer spans in the source and target language for the QA tasks in the appendix. We do not provide translation-based baselines for structured prediction tasks due to an abundance of in-language data and a requirement for annotation projection.

\noindent \textbf{Translate-train multi-task} $\:$ We also experiment with a multi-task version of the translate-train setting where we fine-tune mBERT on the combined translated training data of all languages jointly.

\noindent \textbf{Translate-test} $\:$ Alternatively, we train the English BERT-Large \cite{Devlin2019bert} model on the English training data and evaluate it on test data that we translated from the target language to English using our in-house MT system. 

\noindent \textbf{In-language model} $\:$ For the POS, NER, and TyDiQA-GoldP tasks where target-language training data is available, we fine-tune mBERT on monolingual data in the target language to estimate how useful target language labelled data is compared to labelled data in a source language.

\noindent \textbf{In-language few-shot} $\:$ In many cases, it may be possible to procure a small number of labelled examples in the target language \cite{Eisenschlos2019multifit}. To evaluate the viability of such an approach, we additionally compare against an mBERT model fine-tuned on 1,000 target language examples for the tasks where monolingual training data is available in the target languages.

\begin{table*}[]
\caption{Overall results of baselines across all \task tasks. Translation-based baselines are not meaningful for sentence retrieval. We provide in-language baselines where target language training data is available. Note that for the QA tasks, translate-test performance is not directly comparable to the other scores as a small number of test questions were discarded and alignment is measured on the English data.
}
\resizebox{\textwidth}{!}{%
\begin{tabular}{l c cc cc ccc cc}
\toprule
\multirow{2}{*}{Model} & \multirow{2}{*}{Avg} & \multicolumn{2}{c}{Pair sentence} & \multicolumn{2}{c}{Structured prediction} & \multicolumn{3}{c}{Question answering} & \multicolumn{2}{c}{Sentence retrieval} \\
 & & XNLI & PAWS-X & POS & NER & XQuAD & MLQA & TyDiQA-GoldP & BUCC & Tatoeba \\
\midrule
Metrics & & Acc. & Acc. & F1 & F1 & F1 / EM & F1 / EM & F1 / EM & F1 & Acc. \\ \midrule
\multicolumn{11}{l}{\emph{Cross-lingual zero-shot transfer (models are trained on English data)}} \\ \midrule
mBERT & 59.8 & 65.4 & 81.9  & 71.5 & 62.2 &  64.5 / 49.4 & 61.4 / 44.2 & 59.7 / 43.9 & 56.7 & 38.7\\
XLM & 55.7 & 69.1 & 80.9 & 71.3 & 61.2 &  59.8 / 44.3  & 48.5 / 32.6 & 43.6 / 29.1  & 56.8 & 32.6\\
XLM-R Large & 68.2 & 79.2 & 86.4 & 73.8 & 65.4  & 76.6 / 60.8 & 71.6 / 53.2 & 65.1 / 45.0 & 66.0 & 57.3\\
MMTE & 59.5 & 67.4 & 81.3 & 73.5 & 58.3 & 64.4 / 46.2  & 60.3 /  41.4  & 58.1 / 43.8 & 59.8 & 37.9 \\ \midrule
\multicolumn{11}{l}{\emph{Translate-train (models are trained on English training data translated to the target language)}} \\ \midrule
mBERT & - & 74.6  & 86.3 & - & - & 70.0 / 56.0 & 65.6 / 48.0 & 55.1 / 42.1 & - & -  \\
mBERT, multi-task & - & 75.1  & 88.9 & - & - & 72.4 / 58.3 & 67.6 / 49.8 & 64.2 / 49.3 & - & - \\ \midrule
\multicolumn{11}{l}{\emph{Translate-test (models are trained on English data and evaluated on target language data translated to English)}} \\ \midrule
BERT-large & -  & 76.8  & 84.4 & - & - & 76.3 / 62.1 & 72.9 / 55.3 & 72.1 / 56.0 & - & - \\
\midrule
\multicolumn{11}{l}{\emph{In-language models (models are trained on the target language training data)}} \\ \midrule
mBERT, 1000 examples & - & - & - & 87.6 & 77.9  & - & - & 58.7 / 46.5 & - & - \\
mBERT & - & - & - & 89.8 & 88.3 & - & - & 74.5 / 62.7 & - & - \\
mBERT, multi-task  & - & - & - & 91.5 & 89.1 & - & - & 77.6 / 68.0 & - & - \\
\midrule
Human & - & 92.8 & 97.5 & 97.0 & - & 91.2 / 82.3 & 91.2 / 82.3 & 90.1 / - & - & - \\
\bottomrule
\end{tabular}%
}
\label{tab:main-results}
\end{table*}

\noindent \textbf{In-language multi-task} $\:$ For the tasks where monolingual training data is available, we additionally compare against an mBERT model that is jointly trained on the combined training data of all languages.

\noindent \textbf{Human performance} $\:$ For XNLI, PAWS-X, and XQuAD, we obtain human performance estimates from the English datasets they are derived from, MNLI, PAWS-X, and SQuAD respectively \cite{Nangia2019human_vs_muppet,Zhang2019paws,Rajpurkar2016squad}.\footnote{Performance may differ across languages due to many factors but English performance still serves as a reasonable proxy.} For TyDiQA-GoldP, we use the performance estimate of \citet{Clark2020tydiqa}. For MLQA, as answers are annotated using the same format as SQuAD, we employ the same human performance estimate. For POS tagging, we adopt 97\% as a canonical estimate of human performance based on \citet{manning2011part}. We are not able to obtain human performance estimates for NER as annotations have been automatically generated and for sentence retrieval as identifying a translation among a large number of documents is too time-consuming.

\subsection{Results}

\noindent \textbf{Overall results} $\:$ We show the main results in Table \ref{tab:main-results}. XLM-R is the best-performing zero-shot transfer model and generally improves upon mBERT significantly. The improvement is smaller, however, for the structured prediction tasks. MMTE achieves performance competitive with mBERT on most tasks, with stronger results on XNLI, POS, and BUCC.

If a strong MT system is available, translating the training sets provides improvements over using the same model with zero-shot transfer. Translating the test data provides similar benefits compared to translating the training data and is particularly effective for the more complex QA tasks, while being more expensive during inference time. While using an MT system as a black box leads to strong baselines, the MT system could be further improved in the context of data augmentation.


For the tasks where in-language training data is available, multilingual models trained on in-language data outperform zero-shot transfer models. However, zero-shot transfer models nevertheless outperform multilingual models trained on only 1,000 in-language examples on the complex QA tasks as long as more samples in English are available. For the structured prediction tasks, 1,000 in-language examples enable the model to achieve performance that is similar to being trained on the full labelled dataset, similar to findings for classification \cite{Eisenschlos2019multifit}. Finally, multi-task learning on the Translate-train and In-language setting generally improves upon single language training. 

\noindent \textbf{Cross-lingual transfer gap} $\:$ For a number of representative models, we show the cross-lingual transfer gap, i.e. the difference between the performance on the English test set and all other languages in Table \ref{tab:transfer-gap}.\footnote{This comparison should be taken with a grain of salt, as scores across languages are not directly comparable for the tasks where test sets differ, i.e. POS, NER, MLQA, and TyDiQA-GoldP and differences in scores may not be linearly related.} While powerful models such as XLM-R reduce the gap significantly compared to mBERT for challenging tasks such as XQuAD and MLQA, they do not have the same impact on the syntactic structured prediction tasks. On the classification tasks, the transfer learning gap is lowest, indicating that there may be less headroom for progress on these tasks. The use of MT reduces the gap across all tasks. Overall, a large gap remains for all approaches, which indicates much potential for work on cross-lingual transfer.

\begin{table}[]
\centering
\caption{The cross-lingual transfer gap (lower is better) of different models on \task tasks. The transfer gap is the difference between performance on the English test set and the average performance on the other languages. A transfer gap of 0 indicates perfect cross-lingual transfer. For the QA datasets, we only show EM scores. The average gaps are computed over the sentence classification and QA tasks.}
\resizebox{\columnwidth}{!}{%
\begin{tabular}{l@{~}@{~}c@{~}c@{~}c@{~}c@{~}c@{~}@{~}|c@{~}@{~}|c@{~}c@{~}}
\toprule
Model &  XNLI & PAWS-X  & XQuAD & MLQA & TyDiQA-GoldP & Avg & POS & NER \\
\midrule
mBERT & 16.5 & 14.1 & 25.0 & 27.5 & 22.2 & 21.1 & 25.5 & 23.6 \\
XLM-R & 10.2 & 12.4 & 16.3 & 19.1 & 13.3 & 14.3 & 24.3 & 19.8 \\
Translate-train & 7.3 & 9.0 & 17.6 & 22.2 & 24.2 & 16.1  & - & - \\
Translate-test  & 6.7 & 12.0 & 16.3 & 18.3 & 11.2 & 12.9 & - & -\\
\bottomrule
\end{tabular}%
}
\label{tab:transfer-gap}
\end{table}

\section{Analyses} \label{sec:analyses}

We conduct a series of analyses investigating the limitations of state-of-the-art cross-lingual models.

\begin{figure}[!t]
    \centering
    \includegraphics[width=\linewidth]{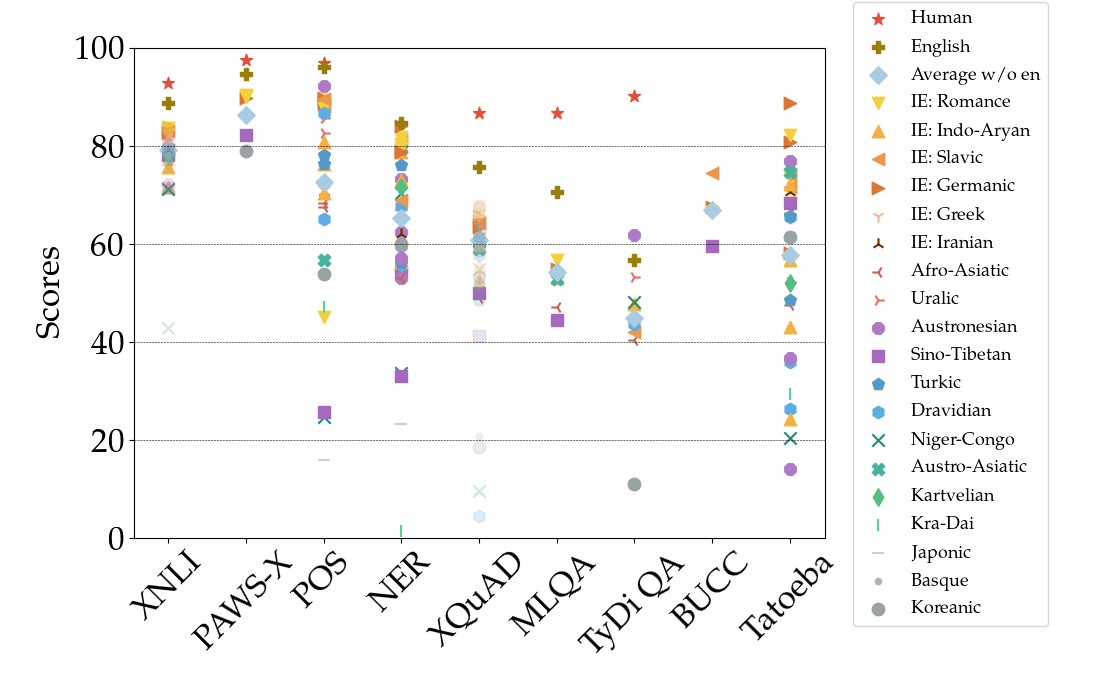}
    \vspace{-0.5cm}
    \caption{An overview of XLM-R's performance on the \task tasks across all languages in each task. We highlight an estimate of human performance, performance on the English test set, the average of all languages excluding English, and the family of each language. Performance on pseudo test sets for XNLI and XQuAD is shown with slightly transparent markers.}
    \label{fig:scores_vs_tasks}
\end{figure}

\noindent \textbf{Best zero-shot model analysis} $\:$ We show the performance of the best zero-shot transfer model, XLM-R Large broken down by task and language in Figure \ref{fig:scores_vs_tasks}. The figure illustrates why it is important to evaluate general-purpose multilingual representations across a diverse range of tasks and languages: On XNLI, probably the most common standard cross-lingual evaluation task, and PAWS-X, scores cluster in a relatively small range---even considering pseudo test sets for XNLI. However, scores for the remaining tasks have significantly wider spread, particularly as we include pseudo test sets. For TyDiQA-GoldP, English performance is lowest in comparison; the high performance on members of the Austronesian and Uralic language families (Indonesian and Finnish) may be due to less complex Wikipedia context passages for these languages. 
Across tasks, we generally observe higher performance on Indo-European languages and lower performance for other language families, particularly for Sino-Tibetan, Japonic, Koreanic, and Niger-Congo languages. Some of these difficulties may be due to tokenisation and an under-representation of ideograms in the joint sentencepiece vocabulary, which has been shown to be important in a cross-lingual model's performance \cite{artetxe2019cross,Conneau2019xlm-r}. We observe similar trends for mBERT, for which we show the same graph in the appendix.

\begin{figure*}[!h]
\centering
\begin{subfigure}{.5\textwidth}
  \centering
  \includegraphics[width=\linewidth]{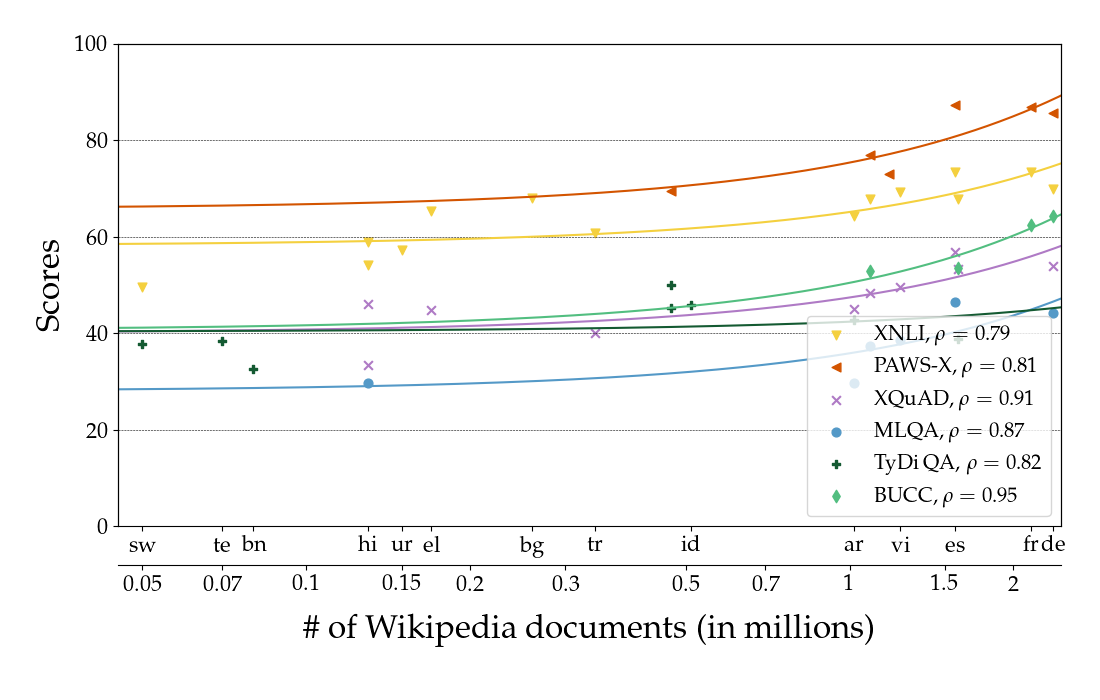}
  \label{fig:sub1}
\end{subfigure}%
\begin{subfigure}{.5\textwidth}
  \centering
  \includegraphics[width=\linewidth]{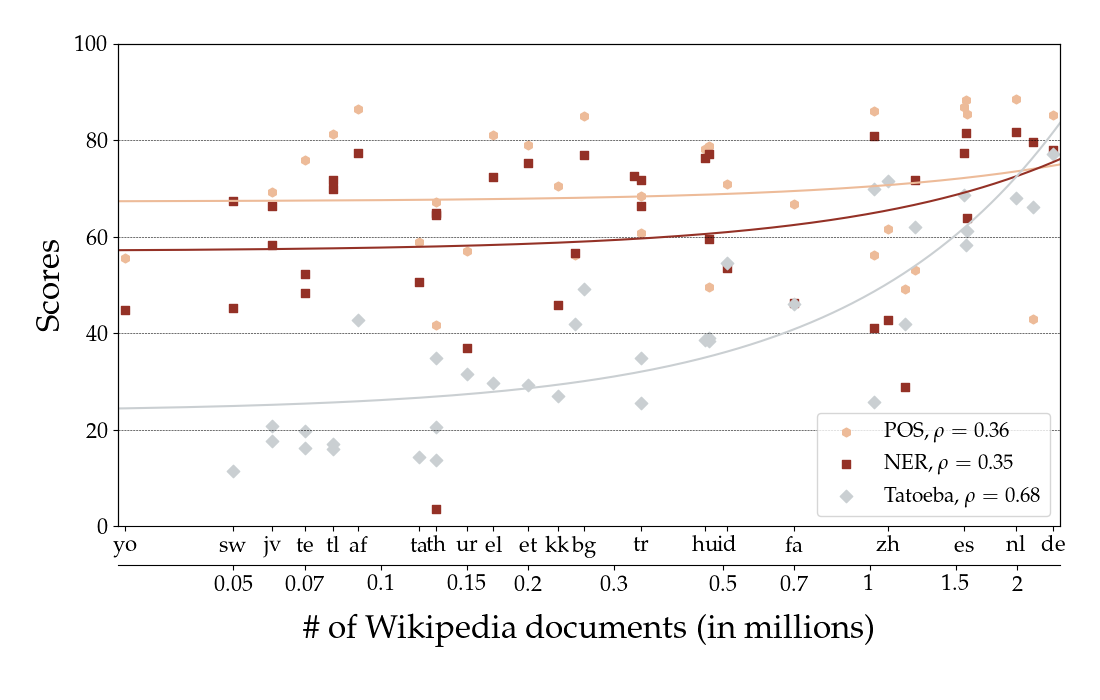}
  \label{fig:sub2}
\end{subfigure}
\vspace{-0.8cm}
\caption{Performance of mBERT across tasks and languages in comparison to the number of Wikipedia articles for each language. We show tasks with a Pearson correlation coefficient $\rho > 0.7$ on the left and others on the right. Numbers across tasks are not directly comparable. We remove the $x$ axis labels of overlapping languages for clarity. We additionally plot the linear fit for each task (curved due to the logarithmic scale of the $x$ axis).}
\label{fig:mbert_correlation_dataset_size}
\end{figure*}

\noindent \textbf{Correlation with pretraining data size} $\:$ We calculate the Pearson correlation coefficient $\rho$ of the model performance and the number of Wikipedia articles (see appendix) in each language and show results in Figure \ref{fig:mbert_correlation_dataset_size}.\footnote{We observe similar correlations when using the number of tokens in Wikipedia instead.} For mBERT, which was pretrained on Wikipedia, we observe a high correlation for most tasks ($\rho \approx 0.8$) except for the structured prediction tasks where $\rho \approx 0.35$.
We observe similar trends for XLM and XLM-R, with lower numbers for XLM-R due to the different pretraining domain (see the appendix). This indicates that current models are not able to fully leverage the information extracted from the pretraining data to transfer to syntactic tasks.

\begin{figure*}[!]
\centering
\begin{subfigure}{.5\textwidth}
  \centering
  \includegraphics[width=\linewidth]{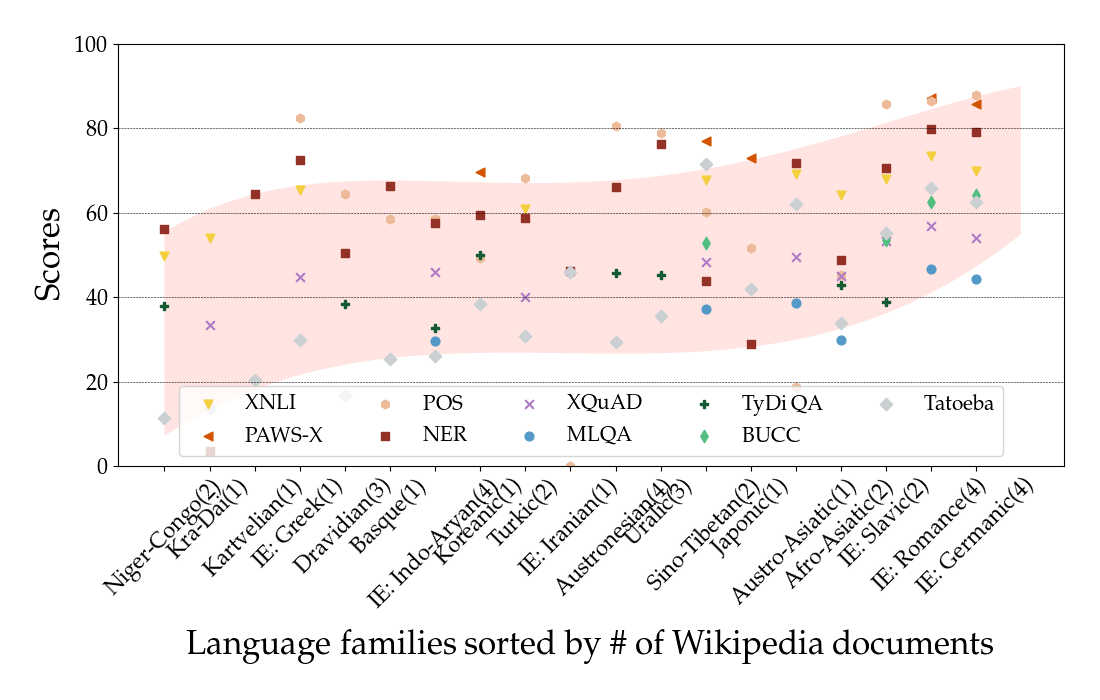}
  \label{fig:lang_families}
\end{subfigure}%
\begin{subfigure}{.5\textwidth}
  \centering
  \includegraphics[width=\linewidth]{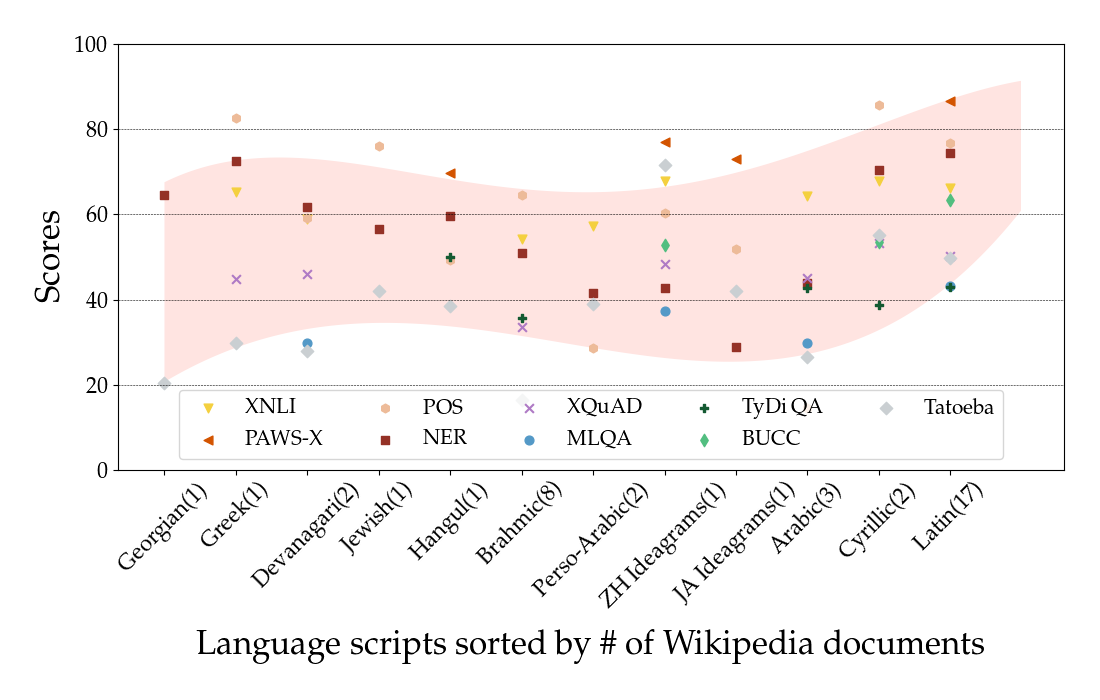}
  \label{fig:lang_scripts}
\end{subfigure}
\vspace{-0.8cm}
\caption{Performance of mBERT across tasks grouped by language families (left) and scripts (right). The number of languages per group is in brackets and the groups are from low-resource to high-resource on the x-axis. We additionally plot the 3rd order polynomial fit for the minimum and maximum values for each group.}
\label{fig:mbert_correlation_lang_family_script}
\end{figure*}

\noindent \textbf{Analysis of language characteristics} $\:$ We analyze results based on different language families and writing scripts in Figure~\ref{fig:mbert_correlation_lang_family_script}. For mBERT, we observe the best transfer performance on branches of the Indo-European language family such as Germanic, Romance and Slavic languages. In contrast, cross-lingual transfer performance on low-resource language families such as Niger-Congo and Kra-Dai is still low. Looking at scripts, we find that the performance on syntactic tasks differs among popular scripts such as Latin and ideogram scripts. For example in the NER task, mBERT performs better on data in Latin script than that in Chinese or Japanese ideograms. This indicates that the current models still have difficulty transferring word-level syntactic information across languages written in different scripts.

\noindent \textbf{Errors across languages} $\:$ For XNLI and XQuAD where the other test sets are translations from English, we analyze whether approaches make the same type of errors in the source and target languages. To this end, we explore whether examples that are correctly and incorrectly predicted in English are correctly predicted in other languages. On the XNLI dev set, mBERT correctly predicts on average 71.8\% of examples that were correctly predicted in English. For examples that were misclassified, the model's performance is about random. On average, predictions on XNLI are consistent between English and another language for 68.3\% of examples. On the XQuAD test set, mBERT correctly predicts around 60\% of examples that were correclty predicted in English and 20\% of examples that were incorrectly predicted. While some of these are plausible spans, more work needs to focus on achieving consistent predictions across languages.

\begin{table}[]
\centering
\caption{Accuracy of mBERT on POS tag trigrams and 4-grams in the target language dev data that appeared and did not appear in the English training data. We show the performance on English, the average across all other languages, and their difference.}
\label{tab:tag-ngrams-mbert}
\resizebox{\columnwidth}{!}{%
\begin{tabular}{l c c c c}
\toprule
 & \begin{tabular}[c]{@{}l@{}}trigram,\\ seen\end{tabular} & \begin{tabular}[c]{@{}l@{}}trigram,\\ unseen\end{tabular} & \begin{tabular}[c]{@{}l@{}}4-gram,\\ seen\end{tabular} & \begin{tabular}[c]{@{}l@{}}4-gram,\\ unseen\end{tabular} \\ \midrule
en & 90.3 & 63.0 & 88.1 & 67.5 \\
avg w/o en & 50.6 & 12.1 & 44.3 & 18.3 \\ \midrule
difference & 39.7 & 50.9 & 43.7 & 49.2 \\ \bottomrule
\end{tabular}%
}
\end{table}

\noindent \textbf{Generalization to unseen tag combinations and entities} $\:$ We analyze possible reasons for the less successful transfer on structured prediction tasks. The Universal Dependencies dataset used for POS tagging uses a common set of 17 POS tags for all languages, so a model is not required to generalize to unseen tags at test time. However, a model may be required to generalize to unseen tag \emph{combinations} at test time, for instance due to differences in word order between languages. We gauge how challenging such generalization is by computing a model's accuracy for POS tag n-grams in the target language dev data that were not seen in the English training data. We calculate values for tag trigrams and 4-grams and show accuracy scores for mBERT in Table \ref{tab:tag-ngrams-mbert}. We observe the largest differences in performance for unseen trigrams and 4-grams, which highlights that existing cross-lingual models struggle to transfer to the syntactic characteristics of other languages. For NER, we estimate how well models generalize to unseen entities at test time. We compute mBERT's accuracy on entities in the target language dev data that were not seen in the English training data. We observe the largest difference between performance on seen and unseen entities for Indonesian and Swahili. Isolating for confounding factors such as entity length, frequency, and Latin script, we find the largest differences in performance for Swahili and Basque. Together, this indicates that the model may struggle to generalize to entities that are more characteristic of the target language. We show the detailed results for both analyses in the appendix.

\section{Conclusions}

As we have highlighted in our analysis, a model's cross-lingual transfer performance varies significantly both between tasks and languages.  \task is a first step towards obtaining a more accurate estimate of a model's cross-lingual generalization ability. While \task is still inherently limited by the data coverage of its constituent tasks for many low-resource languages, \task nevertheless provides significantly broader coverage and more fine-grained analysis tools to encourage research on cross-lingual generalization ability of models. We have released the code for \task and scripts for fine-tuning models on tasks in \task, which should be to catalyze future research.


\section*{Acknowledgements}

We'd like to thank Jon Clark for sharing with us the TyDiQA Gold Passage data and for valuable feedback. We would also like to thank Sam Bowman, Sebastian Goodman, and Tal Linzen for their feedback. JH and GN are sponsored by the Air Force Research Laboratory under agreement number FA8750-19-2-0200. 

\bibliography{icml2020_bib}

\begin{thebibliography}{63}
\providecommand{\natexlab}[1]{#1}
\providecommand{\url}[1]{\texttt{#1}}
\expandafter\ifx\csname urlstyle\endcsname\relax
  \providecommand{\doi}[1]{doi: #1}\else
  \providecommand{\doi}{doi: \begingroup \urlstyle{rm}\Url}\fi

\bibitem[Agi{\'{c}} \& Schluter(2018)Agi{\'{c}} and
  Schluter]{Agic2018baselines}
Agi{\'{c}}, {\v{Z}}. and Schluter, N.
\newblock {Baselines and test data for cross-lingual inference}.
\newblock In \emph{Proceedings of LREC 2018}, 2018.

\bibitem[Aharoni et~al.(2019)Aharoni, Johnson, and Firat]{aharoni2019massively}
Aharoni, R., Johnson, M., and Firat, O.
\newblock {Massively Multilingual Neural Machine Translation}.
\newblock In \emph{Proceedings of NAACL 2019}, 2019.

\bibitem[Arivazhagan et~al.(2019)Arivazhagan, Bapna, Firat, Lepikhin, Johnson,
  Krikun, Chen, Cao, Foster, Cherry, Macherey, Chen, and
  Wu]{Arivazhagan2019massively_multilingual}
Arivazhagan, N., Bapna, A., Firat, O., Lepikhin, D., Johnson, M., Krikun, M.,
  Chen, M.~X., Cao, Y., Foster, G., Cherry, C., Macherey, W., Chen, Z., and Wu,
  Y.
\newblock {Massively Multilingual Neural Machine Translation in the Wild:
  Findings and Challenges}.
\newblock \emph{arXiv preprint arXiv:1907.05019}, 2019.

\bibitem[Artetxe \& Schwenk(2019)Artetxe and Schwenk]{Artetxe2019massively}
Artetxe, M. and Schwenk, H.
\newblock {Massively Multilingual Sentence Embeddings for Zero-Shot
  Cross-Lingual Transfer and Beyond}.
\newblock \emph{Transactions of the ACL 2019}, 2019.

\bibitem[Artetxe et~al.(2017)Artetxe, Labaka, and Agirre]{artetxe2017learning}
Artetxe, M., Labaka, G., and Agirre, E.
\newblock Learning bilingual word embeddings with (almost) no bilingual data.
\newblock In \emph{Proceedings of ACL 2017}, pp.\  451--462, 2017.

\bibitem[Artetxe et~al.(2018)Artetxe, Labaka, and Agirre]{artetxe2018robust}
Artetxe, M., Labaka, G., and Agirre, E.
\newblock A robust self-learning method for fully unsupervised cross-lingual
  mappings of word embeddings.
\newblock In \emph{Proceedings of ACL 2018}, pp.\  789--798, 2018.

\bibitem[Artetxe et~al.(2020)Artetxe, Ruder, and Yogatama]{artetxe2019cross}
Artetxe, M., Ruder, S., and Yogatama, D.
\newblock {On the Cross-lingual Transferability of Monolingual
  Representations}.
\newblock In \emph{Proceedings of ACL 2020}, 2020.

\bibitem[Barnes et~al.(2018)Barnes, Klinger, and Schulte~im
  Walde]{barnes2018bilingual}
Barnes, J., Klinger, R., and Schulte~im Walde, S.
\newblock Bilingual sentiment embeddings: Joint projection of sentiment across
  languages.
\newblock In \emph{Proceedings of ACL 2018}, pp.\  2483--2493, Melbourne,
  Australia, 2018. Association for Computational Linguistics.

\bibitem[Clark et~al.(2020)Clark, Choi, Collins, Garrette, Kwiatkowski,
  Nikolaev, and Palomaki]{Clark2020tydiqa}
Clark, J.~H., Choi, E., Collins, M., Garrette, D., Kwiatkowski, T., Nikolaev,
  V., and Palomaki, J.
\newblock {TyDi QA: A Benchmark for Information-Seeking Question Answering in
  Typologically Diverse Languages}.
\newblock In \emph{Transactions of the Association of Computational
  Linguistics}, 2020.

\bibitem[Conneau et~al.(2018{\natexlab{a}})Conneau, Lample, Ranzato, Denoyer,
  and J{\'{e}}gou]{conneau2018word}
Conneau, A., Lample, G., Ranzato, M., Denoyer, L., and J{\'{e}}gou, H.
\newblock Word translation without parallel data.
\newblock In \emph{Proceedings of ICLR 2018}, 2018{\natexlab{a}}.

\bibitem[Conneau et~al.(2018{\natexlab{b}})Conneau, Rinott, Lample, Williams,
  Bowman, Schwenk, and Stoyanov]{Conneau2018xnli}
Conneau, A., Rinott, R., Lample, G., Williams, A., Bowman, S., Schwenk, H., and
  Stoyanov, V.
\newblock {XNLI}: Evaluating cross-lingual sentence representations.
\newblock In \emph{Proceedings of EMNLP 2018}, pp.\  2475--2485,
  2018{\natexlab{b}}.

\bibitem[Conneau et~al.(2020)Conneau, Khandelwal, Goyal, Chaudhary, Wenzek,
  Guzm{\'{a}}n, Grave, Ott, Zettlemoyer, and Stoyanov]{Conneau2019xlm-r}
Conneau, A., Khandelwal, K., Goyal, N., Chaudhary, V., Wenzek, G.,
  Guzm{\'{a}}n, F., Grave, E., Ott, M., Zettlemoyer, L., and Stoyanov, V.
\newblock {Unsupervised Cross-lingual Representation Learning at Scale}.
\newblock In \emph{Proceedings of ACL 2020}, 2020.

\bibitem[Czarnowska et~al.(2019)Czarnowska, Ruder, Grave, Cotterell, and
  Copestake]{czarnowska2019dont}
Czarnowska, P., Ruder, S., Grave, E., Cotterell, R., and Copestake, A.
\newblock Don{'}t forget the long tail! a comprehensive analysis of
  morphological generalization in bilingual lexicon induction.
\newblock In \emph{Proceedings of EMNLP 2019}, pp.\  973--982, 2019.

\bibitem[Devlin et~al.(2019)Devlin, Chang, Lee, and Toutanova]{Devlin2019bert}
Devlin, J., Chang, M.-W., Lee, K., and Toutanova, K.
\newblock {BERT}: Pre-training of deep bidirectional transformers for language
  understanding.
\newblock In \emph{Proceedings of NAACL 2019}, 2019.

\bibitem[Eisenschlos et~al.(2019)Eisenschlos, Ruder, Czapla, Kadras, Gugger,
  and Howard]{Eisenschlos2019multifit}
Eisenschlos, J., Ruder, S., Czapla, P., Kadras, M., Gugger, S., and Howard, J.
\newblock {MultiFiT: Efficient Multi-lingual Language Model Fine-tuning}.
\newblock In \emph{Proceedings of EMNLP 2019}, 2019.

\bibitem[Eriguchi et~al.(2018)Eriguchi, Johnson, Firat, Kazawa, and
  Macherey]{eriguchi2018zero}
Eriguchi, A., Johnson, M., Firat, O., Kazawa, H., and Macherey, W.
\newblock Zero-shot cross-lingual classification using multilingual neural
  machine translation.
\newblock \emph{arXiv preprint arXiv:1809.04686}, 2018.

\bibitem[Faruqui \& Dyer(2014)Faruqui and Dyer]{faruqui2014improving}
Faruqui, M. and Dyer, C.
\newblock Improving vector space word representations using multilingual
  correlation.
\newblock In \emph{Proceedings of EACL 2014}, pp.\  462--471, 2014.

\bibitem[Glava{\v{s}} et~al.(2019)Glava{\v{s}}, Litschko, Ruder, and
  Vuli{\'{c}}]{Glavas2019}
Glava{\v{s}}, G., Litschko, R., Ruder, S., and Vuli{\'{c}}, I.
\newblock {How to (Properly) Evaluate Cross-Lingual Word Embeddings: On Strong
  Baselines, Comparative Analyses, and Some Misconceptions}.
\newblock In \emph{Proceedings of ACL 2019}, 2019.

\bibitem[Gouws et~al.(2015)Gouws, Bengio, and Corrado]{gouws2015bilbowa}
Gouws, S., Bengio, Y., and Corrado, G.
\newblock {BilBOWA}: Fast bilingual distributed representations without word
  alignments.
\newblock In \emph{Proceedings of ICML 2015}, pp.\  748--756, 2015.

\bibitem[Gururangan et~al.(2018)Gururangan, Swayamdipta, Levy, Schwartz,
  Bowman, and Smith]{Gururangan2018}
Gururangan, S., Swayamdipta, S., Levy, O., Schwartz, R., Bowman, S.~R., and
  Smith, N.~A.
\newblock {Annotation Artifacts in Natural Language Inference Data}.
\newblock In \emph{Proceedings of NAACL-HLT 2018}, 2018.

\bibitem[Guzm{\'{a}}n et~al.(2019)Guzm{\'{a}}n, Chen, Ott, Pino, Lample, Koehn,
  Chaudhary, and Ranzato]{Guzman2019}
Guzm{\'{a}}n, F., Chen, P.-J., Ott, M., Pino, J., Lample, G., Koehn, P.,
  Chaudhary, V., and Ranzato, M.
\newblock {The FLoRes Evaluation Datasets for Low-Resource Machine Translation:
  Nepali-English and Sinhala-English}.
\newblock In \emph{Proceedings of EMNLP 2019}, pp.\  6100--6113, 2019.

\bibitem[Howard \& Ruder(2018)Howard and Ruder]{howard2018universal}
Howard, J. and Ruder, S.
\newblock Universal language model fine-tuning for text classification.
\newblock In \emph{Proceedings of ACL 2018}, pp.\  328--339, 2018.

\bibitem[Hsu et~al.(2019)Hsu, Liu, and Lee]{Hsu2019zero-shot}
Hsu, T.-y., Liu, C.-l., and Lee, H.-y.
\newblock {Zero-shot Reading Comprehension by Cross-lingual Transfer Learning
  with Multi-lingual Language Representation Model}.
\newblock In \emph{Proceedings of EMNLP 2019}, pp.\  5935--5942, 2019.

\bibitem[Kementchedjhieva et~al.(2019)Kementchedjhieva, Hartmann, and
  S{\o}gaard]{kementchedjhieva2019lost}
Kementchedjhieva, Y., Hartmann, M., and S{\o}gaard, A.
\newblock Lost in evaluation: Misleading benchmarks for bilingual dictionary
  induction.
\newblock In \emph{Proceedings of EMNLP 2019}, pp.\  3327--3332, 2019.

\bibitem[Klementiev et~al.(2012)Klementiev, Titov, and
  Bhattarai]{Klementiev2012inducing}
Klementiev, A., Titov, I., and Bhattarai, B.
\newblock {Inducing Crosslingual Distributed Representations of Words}.
\newblock In \emph{Proceedings of COLING 2012}, 2012.

\bibitem[Koppel \& Ordan()Koppel and Ordan]{koppel2011translationese}
Koppel, M. and Ordan, N.
\newblock Translationese and its dialects.
\newblock In \emph{Proceedings of ACL 2011, pages={1318--1326}, year={2011},
  organization={Association for Computational Linguistics}}.

\bibitem[Lample \& Conneau(2019)Lample and Conneau]{Lample2019xlm}
Lample, G. and Conneau, A.
\newblock {Cross-lingual Language Model Pretraining}.
\newblock In \emph{Proceedings of NeurIPS 2019}, 2019.

\bibitem[Lee et~al.(2018)Lee, Yoon, Park, and Hwang]{Lee2018semi-supervised}
Lee, K., Yoon, K., Park, S., and Hwang, S.~W.
\newblock {Semi-supervised training data generation for multilingual question
  answering}.
\newblock In \emph{Proceedings of LREC 2018}, pp.\  2758--2762, 2018.

\bibitem[Lewis et~al.(2019)Lewis, Oğuz, Rinott, Riedel, and
  Schwenk]{Lewis2019mlqa}
Lewis, P., Oğuz, B., Rinott, R., Riedel, S., and Schwenk, H.
\newblock {MLQA: Evaluating Cross-lingual Extractive Question Answering}.
\newblock \emph{arXiv preprint arXiv:1910.07475}, 2019.

\bibitem[Lin et~al.(2019)Lin, Chen, Lee, Li, Zhang, Xia, Rijhwani, He, Zhang,
  Ma, Anastasopoulos, Littell, and Neubig]{Lin2019transfer_languages}
Lin, Y.-H., Chen, C.-Y., Lee, J., Li, Z., Zhang, Y., Xia, M., Rijhwani, S., He,
  J., Zhang, Z., Ma, X., Anastasopoulos, A., Littell, P., and Neubig, G.
\newblock {Choosing Transfer Languages for Cross-Lingual Learning}.
\newblock In \emph{Proceedings of ACL 2019}, 2019.

\bibitem[Luong et~al.(2015)Luong, Pham, and Manning]{luong2015bilingual}
Luong, T., Pham, H., and Manning, C.~D.
\newblock Bilingual word representations with monolingual quality in mind.
\newblock In \emph{Proceedings of the 1st Workshop on Vector Space Modeling for
  Natural Language Processing}, pp.\  151--159, 2015.

\bibitem[Manning(2011)]{manning2011part}
Manning, C.~D.
\newblock Part-of-speech tagging from 97\% to 100\%: is it time for some
  linguistics?
\newblock In \emph{International conference on intelligent text processing and
  computational linguistics}, pp.\  171--189. Springer, 2011.

\bibitem[McCann et~al.(2017)McCann, Bradbury, Xiong, and
  Socher]{mccann2017learned}
McCann, B., Bradbury, J., Xiong, C., and Socher, R.
\newblock Learned in translation: Contextualized word vectors.
\newblock In \emph{Proceedings of NIPS 2017}, pp.\  6294--6305, 2017.

\bibitem[McDonald et~al.(2011)McDonald, Petrov, and
  Hall]{McDonald2011delexicalized}
McDonald, R., Petrov, S., and Hall, K.
\newblock {Multi-source transfer of delexicalized dependency parsers}.
\newblock In \emph{Proceedings of EMNLP 2011}, pp.\  62--72, 2011.

\bibitem[Mikolov et~al.(2013)Mikolov, Le, and Sutskever]{mikolov2013exploiting}
Mikolov, T., Le, Q.~V., and Sutskever, I.
\newblock Exploiting similarities among languages for machine translation.
\newblock \emph{arXiv preprint arXiv:1309.4168}, 2013.

\bibitem[Mohammad et~al.(2016)Mohammad, Salameh, and
  Kiritchenko]{mohammad2016translation}
Mohammad, S.~M., Salameh, M., and Kiritchenko, S.
\newblock How translation alters sentiment.
\newblock \emph{Journal of Artificial Intelligence Research}, 55:\penalty0
  95--130, 2016.

\bibitem[Nangia \& Bowman(2019)Nangia and Bowman]{Nangia2019human_vs_muppet}
Nangia, N. and Bowman, S.~R.
\newblock {Human vs. Muppet: A Conservative Estimate of Human Performance on
  the GLUE Benchmark}.
\newblock In \emph{Proceedings of ACL 2019}, pp.\  4566--4575, 2019.

\bibitem[Nivre et~al.(2018)Nivre, Abrams, Agi{\'c}, Ahrenberg, Antonsen,
  Aranzabe, Arutie, Asahara, Ateyah, Attia, et~al.]{nivre2018universal}
Nivre, J., Abrams, M., Agi{\'c}, {\v{Z}}., Ahrenberg, L., Antonsen, L.,
  Aranzabe, M.~J., Arutie, G., Asahara, M., Ateyah, L., Attia, M., et~al.
\newblock Universal dependencies 2.2.
\newblock 2018.

\bibitem[Pan et~al.(2017)Pan, Zhang, May, Nothman, Knight, and Ji]{Pan2017}
Pan, X., Zhang, B., May, J., Nothman, J., Knight, K., and Ji, H.
\newblock {Cross-lingual name tagging and linking for 282 languages}.
\newblock In \emph{Proceedings of ACL 2017}, pp.\  1946--1958, 2017.

\bibitem[Peters et~al.(2018)Peters, Neumann, Iyyer, Gardner, Clark, Lee, and
  Zettlemoyer]{peters2018deep}
Peters, M., Neumann, M., Iyyer, M., Gardner, M., Clark, C., Lee, K., and
  Zettlemoyer, L.
\newblock Deep contextualized word representations.
\newblock In \emph{Proceedings of NAACL 2018}, pp.\  2227--2237, 2018.

\bibitem[Pires et~al.(2019)Pires, Schlinger, and Garrette]{Pires2019}
Pires, T., Schlinger, E., and Garrette, D.
\newblock {How multilingual is Multilingual BERT?}
\newblock In \emph{Proceedings of ACL 2019}, 2019.

\bibitem[Popovi{\'c}(2015)]{popovic-2015-chrf}
Popovi{\'c}, M.
\newblock chr{F}: character n-gram f-score for automatic {MT} evaluation.
\newblock In \emph{Proceedings of the Tenth Workshop on Statistical Machine
  Translation}, pp.\  392--395, Lisbon, Portugal, 2015.

\bibitem[Rahimi et~al.(2019)Rahimi, Li, and Cohn]{Rahimi2019}
Rahimi, A., Li, Y., and Cohn, T.
\newblock {Massively Multilingual Transfer for NER}.
\newblock In \emph{Proceedings of ACL 2019}, 2019.

\bibitem[Rajpurkar et~al.(2016)Rajpurkar, Zhang, Lopyrev, and
  Liang]{Rajpurkar2016squad}
Rajpurkar, P., Zhang, J., Lopyrev, K., and Liang, P.
\newblock {SQuAD: 100,000+ Questions for Machine Comprehension of Text}.
\newblock In \emph{Proceedings of EMNLP 2016}, 2016.

\bibitem[Ruder et~al.(2019)Ruder, Vuli{\'{c}}, and S{\o}gaard]{Ruder2019survey}
Ruder, S., Vuli{\'{c}}, I., and S{\o}gaard, A.
\newblock {A Survey of Cross-lingual Word Embedding Models}.
\newblock \emph{Journal of Artificial Intelligence Research}, 65:\penalty0
  569--631, 2019.

\bibitem[Schuster et~al.(2019)Schuster, Ram, Barzilay, and
  Globerson]{Schuster2019}
Schuster, T., Ram, O., Barzilay, R., and Globerson, A.
\newblock {Cross-Lingual Alignment of Contextual Word Embeddings, with
  Applications to Zero-shot Dependency Parsing}.
\newblock In \emph{Proceedings of NAACL 2019}, 2019.

\bibitem[Schwenk \& Li(2018)Schwenk and Li]{Schwenk2018mldoc}
Schwenk, H. and Li, X.
\newblock {A Corpus for Multilingual Document Classification in Eight
  Languages}.
\newblock In \emph{Proceedings of LREC 2018}, 2018.

\bibitem[Siddhant et~al.(2020)Siddhant, Johnson, Tsai, Arivazhagan, Riesa,
  Bapna, Firat, and Raman]{Siddhant2019evaluating}
Siddhant, A., Johnson, M., Tsai, H., Arivazhagan, N., Riesa, J., Bapna, A.,
  Firat, O., and Raman, K.
\newblock {Evaluating the Cross-Lingual Effectiveness of Massively Multilingual
  Neural Machine Translation}.
\newblock In \emph{Proceedings of AAAI 2020}, 2020.

\bibitem[Smith et~al.(2016)Smith, Giorgi, Solanki, Eichstaedt, Schwartz,
  Abdul-Mageed, Buffone, and Ungar]{smith2016does}
Smith, L., Giorgi, S., Solanki, R., Eichstaedt, J., Schwartz, H.~A.,
  Abdul-Mageed, M., Buffone, A., and Ungar, L.
\newblock Does ‘well-being’ translate on twitter?
\newblock In \emph{Proceedings of EMNLP 2016}, pp.\  2042--2047, 2016.

\bibitem[Snyder et~al.(2009)Snyder, Naseem, and
  Barzilay]{Snyder2009unsupervised}
Snyder, B., Naseem, T., and Barzilay, R.
\newblock {Unsupervised multilingual grammar induction}.
\newblock In \emph{Proceedings of ACL 2009}, pp.\  73--81, 2009.

\bibitem[T{\"{a}}ckstr{\"{o}}m et~al.(2013)T{\"{a}}ckstr{\"{o}}m, Das, Petrov,
  McDonald, and Nivre]{Tackstrom2013token_and_type}
T{\"{a}}ckstr{\"{o}}m, O., Das, D., Petrov, S., McDonald, R., and Nivre, J.
\newblock {Token and Type Constraints for Cross-Lingual Part-of-Speech
  Tagging}.
\newblock In \emph{Transactions of the Association for Computational
  Linguistics}, 2013.

\bibitem[Vaswani et~al.(2017)Vaswani, Shazeer, Parmar, Uszkoreit, Jones, Gomez,
  Kaiser, and Polosukhin]{Vaswani2017attention}
Vaswani, A., Shazeer, N., Parmar, N., Uszkoreit, J., Jones, L., Gomez, A.~N.,
  Kaiser, {\L}., and Polosukhin, I.
\newblock {Attention Is All You Need}.
\newblock In \emph{Proceedings of NIPS 2017}, 2017.

\bibitem[Vuli{\'{c}} et~al.(2019)Vuli{\'{c}}, Glava{\v{s}}, Reichart, and
  Korhonen]{Vulic2019}
Vuli{\'{c}}, I., Glava{\v{s}}, G., Reichart, R., and Korhonen, A.
\newblock {Do We Really Need Fully Unsupervised Cross-Lingual Embeddings?}
\newblock In \emph{Proceedings of EMNLP 2019}, 2019.

\bibitem[Wang et~al.(2019{\natexlab{a}})Wang, Pruksachatkun, Nangia, Singh,
  Michael, Hill, Levy, and Bowman]{wang2019superglue}
Wang, A., Pruksachatkun, Y., Nangia, N., Singh, A., Michael, J., Hill, F.,
  Levy, O., and Bowman, S.~R.
\newblock Superglue: A stickier benchmark for general-purpose language
  understanding systems.
\newblock In \emph{Proceedings of NeurIPS 2019}, 2019{\natexlab{a}}.

\bibitem[Wang et~al.(2019{\natexlab{b}})Wang, Singh, Michael, Hill, Levy, and
  Bowman]{Wang2019glue}
Wang, A., Singh, A., Michael, J., Hill, F., Levy, O., and Bowman, S.~R.
\newblock {GLUE: A Multi-Task Benchmark and Analysis Platform for Natural
  Language Understanding}.
\newblock In \emph{Proceedings of ICLR 2019}, 2019{\natexlab{b}}.

\bibitem[Williams et~al.(2018)Williams, Nangia, and
  Bowman]{Williams2018multinli}
Williams, A., Nangia, N., and Bowman, S.~R.
\newblock {A Broad-Coverage Challenge Corpus for Sentence Understanding through
  Inference}.
\newblock In \emph{Proceedings of NAACL-HLT 2018}, 2018.

\bibitem[Wolf et~al.(2019)Wolf, Debut, Sanh, Chaumond, Delangue, Moi, Cistac,
  Rault, Louf, Funtowicz, et~al.]{wolf2019huggingface}
Wolf, T., Debut, L., Sanh, V., Chaumond, J., Delangue, C., Moi, A., Cistac, P.,
  Rault, T., Louf, R., Funtowicz, M., et~al.
\newblock Huggingface's transformers: State-of-the-art natural language
  processing.
\newblock \emph{arXiv preprint arXiv:1910.03771}, 2019.

\bibitem[Wu \& Dredze(2019)Wu and Dredze]{Wu2019}
Wu, S. and Dredze, M.
\newblock {Beto, Bentz, Becas: The Surprising Cross-Lingual Effectiveness of
  BERT}.
\newblock In \emph{Proceedings of EMNLP 2019}, 2019.

\bibitem[Yang et~al.(2019)Yang, Zhang, Tar, and Baldridge]{Yang2019paws-x}
Yang, Y., Zhang, Y., Tar, C., and Baldridge, J.
\newblock {PAWS-X}: A cross-lingual adversarial dataset for paraphrase
  identification.
\newblock In \emph{Proceedings of EMNLP 2019}, pp.\  3685--3690, 2019.

\bibitem[Zhang et~al.(2017)Zhang, Liu, Luan, and Sun]{zhang2017earth}
Zhang, M., Liu, Y., Luan, H., and Sun, M.
\newblock Earth mover's distance minimization for unsupervised bilingual
  lexicon induction.
\newblock In \emph{Proceedings of EMNLP 2017}, pp.\  1934--1945, 2017.

\bibitem[Zhang et~al.(2019)Zhang, Baldridge, and He]{Zhang2019paws}
Zhang, Y., Baldridge, J., and He, L.
\newblock {PAWS}: Paraphrase adversaries from word scrambling.
\newblock In \emph{Proceedings of NAACL 2019}, pp.\  1298--1308, 2019.

\bibitem[Zweigenbaum et~al.(2017)Zweigenbaum, Sharoff, and
  Rapp]{zweigenbaum2017overview}
Zweigenbaum, P., Sharoff, S., and Rapp, R.
\newblock Overview of the second bucc shared task: Spotting parallel sentences
  in comparable corpora.
\newblock In \emph{Proceedings of the 10th Workshop on Building and Using
  Comparable Corpora}, pp.\  60--67, 2017.

\bibitem[Zweigenbaum et~al.(2018)Zweigenbaum, Sharoff, and
  Rapp]{zweigenbaum2018overview}
Zweigenbaum, P., Sharoff, S., and Rapp, R.
\newblock Overview of the third bucc shared task: Spotting parallel sentences
  in comparable corpora.
\newblock In \emph{Proceedings of 11th Workshop on Building and Using
  Comparable Corpora}, pp.\  39--42, 2018.

\end{thebibliography}
\bibliographystyle{icml2020}

\newpage

\appendix

\section{Languages}
We show a detailed overview of languages in the cross-lingual benchmark including interesting typological differences in Table \ref{tab:languages}. Wikipedia information is taken from Wikipedia\footnote{\url{https://meta.wikimedia.org/wiki/List_of_Wikipedias}} and linguistic information from WALS Online\footnote{\url{https://wals.info/languoid}}. \task includes members of the Afro-Asiatic, Austro-Asiatic, Austronesian, Dravidian, Indo-European, Japonic, Kartvelian, Kra-Dai, Niger-Congo, Sino-Tibetan, Turkic, and Uralic language families as well as of two isolates, Basque and Korean.

\begin{table*}[]
\centering
\caption{Statistics about languages in the cross-lingual benchmark. Languages belong to 12 language families and two isolates, with Indo-European (IE) having the most members. Diacritics / special characters: Language adds diacritics (additional symbols to letters). Compounding: Language makes extensive use of word compounds. Bound words / clitics: Function words attach to other words. Inflection: Words are inflected to represent grammatical meaning (e.g.~case marking). Derivation: A single token can represent entire phrases or sentences.}
\label{tab:languages}
\resizebox{\textwidth}{!}{%
\begin{tabular}{l c c l l ccccc c}
\toprule
Language & \begin{tabular}[c]{@{}l@{}}ISO\\ 639-1\\ code\end{tabular} & \begin{tabular}[c]{@{}l@{}}\# Wikipedia\\ articles (in\\ millions)\end{tabular} & Script & \begin{tabular}[c]{@{}l@{}}Language\\ family\end{tabular} & \begin{tabular}[c]{@{}l@{}}Diacritics /\\ special\\ characters\end{tabular} & \begin{tabular}[c]{@{}l@{}}Extensive\\ compound-\\ ing\end{tabular} & \begin{tabular}[c]{@{}l@{}}Bound\\ words /\\ clitics\end{tabular} & \begin{tabular}[c]{@{}l@{}}Inflec-\\ tion\end{tabular} & \begin{tabular}[c]{@{}l@{}}Deriva-\\ tion\end{tabular} & \begin{tabular}[c]{@{}l@{}}\# datasets\\ with\\ language\end{tabular} \\ \midrule
Afrikaans & af & 0.09 & Latin & IE: Germanic &  & X &  &  &  & 3 \\
Arabic & ar & 1.02 & Arabic & Afro-Asiatic & X &  & X & X &  & 7 \\
Basque & eu & 0.34 & Latin & Basque & X &  & X & X & X & 3 \\
Bengali & bn & 0.08 & Brahmic & IE: Indo-Aryan & X & X & X & X & X & 3 \\
Bulgarian & bg & 0.26 & Cyrillic & IE: Slavic & X &  & X & X &  & 4 \\
Burmese & my & 0.05 & Brahmic & Sino-Tibetan & X & X &  &  &  & 1 \\
Dutch & nl & 1.99 & Latin & IE: Germanic &  & X &  &  &  & 3 \\
English & en & 5.98 & Latin & IE: Germanic &  &  &  &  &  & 9 \\
Estonian & et & 0.20 & Latin & Uralic & X & X &  & X & X & 3 \\
Finnish & fi & 0.47 & Latin & Uralic &  &  &  & X & X & 3 \\
French & fr & 2.16 & Latin & IE: Romance & X &  & X &  &  & 6 \\
Georgian & ka & 0.13 & Georgian & Kartvelian &  &  &  & X & X & 2 \\
German & de & 2.37 & Latin & IE: Germanic &  & X &  & X &  & 8 \\
Greek & el & 0.17 & Greek & IE: Greek & X & X &  & X &  & 5 \\
Hebrew & he & 0.25 & Hebrew & Afro-Asiatic &  &  &  & X &  & 3 \\
Hindi & hi & 0.13 & Devanagari & IE: Indo-Aryan & X & X & X & X & X & 6 \\
Hungarian & hu & 0.46 & Latin & Uralic & X & X &  & X & X & 4 \\
Indonesian & id & 0.51 & Latin & Austronesian &  &  & X & X & X & 4 \\
Italian & it & 1.57 & Latin & IE: Romance & X &  & X &  &  & 3 \\
Japanese & ja & 1.18 & Ideograms & Japonic &  &  & X & X &  & 4 \\
Javanese & jv & 0.06 & Brahmic & Austronesian & X &  & X &  &  & 1 \\
Kazakh & kk & 0.23 & Arabic & Turkic & X &  &  & X & X & 1 \\
Korean & ko & 0.47 & Hangul & Koreanic &  & X &  & X & X & 5 \\
Malay & ms & 0.33 & Latin & Austronesian &  &  & X & X &  & 2 \\
Malayalam & ml & 0.07 & Brahmic & Dravidian & X & X & X & X &  & 2 \\
Mandarin & zh & 1.09 & Chinese ideograms & Sino-Tibetan &  & X &  &  &  & 8 \\
Marathi & mr & 0.06 & Devanagari & IE: Indo-Aryan &  &  & X & X &  & 3 \\
Persian & fa & 0.70 & Perso-Arabic & IE: Iranian &  & X &  &  &  & 2 \\
Portuguese & pt & 1.02 & Latin & IE: Romance & X &  & X &  &  & 3 \\
Russian & ru & 1.58 & Cyrillic & IE: Slavic &  &  &  & X &  & 7 \\
Spanish & es & 1.56 & Latin & IE: Romance & X &  & X &  &  & 7 \\
Swahili & sw & 0.05 & Latin & Niger-Congo &  &  & X & X & X & 3 \\
Tagalog & tl & 0.08 & Brahmic & Austronesian & X &  & X & X &  & 1 \\
Tamil & ta & 0.12 & Brahmic & Dravidian & X & X & X & X & X & 3 \\
Telugu & te & 0.07 & Brahmic & Dravidian & X & X & X & X & X & 4 \\
Thai & th & 0.13 & Brahmic & Kra-Dai & X &  &  &  &  & 4 \\
Turkish & tr & 0.34 & Latin & Turkic & X & X &  & X & X & 5 \\
Urdu & ur & 0.15 & Perso-Arabic & IE: Indo-Aryan & X & X & X & X & X & 4 \\
Vietnamese & vi & 1.24 & Latin & Austro-Asiatic & X &  &  &  &  & 6 \\
Yoruba & yo & 0.03 & Arabic & Niger-Congo & X &  &  &  &  & 1 \\
\bottomrule
\end{tabular}%
}
\end{table*}

\section{Hyper-parameters} \label{app:hyper-parameters}
Table~\ref{tab:params} summarizes the hyper-parameters of baseline and state-of-the-art models. We refer to XLM-100 as XLM, and XLM-R-large as XLM-R in our paper to simplify the notation.

\noindent \textbf{mBERT} $\:$ We use the cased version, which covers 104 languages, has 12 layers, 768 hidden units per layer, 12 attention heads, a 110k shared WordPiece vocabulary, and 110M parameters.\footnote{\url{https://github.com/google-research/bert/blob/master/multilingual.md}} The model was trained using Wikipedia data in all 104 languages, oversampling low-resource languages with an exponential smoothing factor of 0.7. We generally fine-tune mBERT for two epochs, with a training batch size of 32 and a learning rate of 2e-5. For training BERT models on the QA tasks, we use the original BERT codebase. For all other tasks, we use the Transformers library \cite{wolf2019huggingface}.

\noindent \textbf{XLM and XLM-R} $\:$ We use the XLM and XLM-R Large versions that cover 100 languages, use a 200k shared BPE vocabulary, and that have been trained with masked language modelling.\footnote{\url{https://github.com/facebookresearch/XLM}} We fine-tune both for two epochs with a learning rate of 3e-5 and an effective batch size of 16. In contrast to XLM, XLM-R does not use language embeddings. We use the Transformers library for training XLM and XLM-R models on all tasks. 

\begin{table}[]
\caption{Hyper-parameters of baseline and state-of-the-art models. We do not use XLM-15 and XLM-R-Base in our experiments.}
\label{tab:params}
\resizebox{\columnwidth}{!}{%
\begin{tabular}{l cccc}
\toprule
Model       & Parameters & Langs & Vocab size & Layers \\ \midrule
BERT-large  &  364,353,862 & 1    &  28,996      & 24 \\
mBERT       & 178,566,653  & 104   & 119,547     & 12     \\ 
MMTE       & 191,733,123  & 103   & 64,000     & 6     \\ 
XLM-15      & 346,351,384  & 15    & 95,000      & 12     \\ 
XLM-100     & 827,696,960  & 100   & 200,000     & 12     \\ 
XLM-R-Base  & 470,295,954  & 100   & 250,002     & 12     \\ 
XLM-R-Large & 816,143,506  & 100   & 250,002     & 24     \\ 
\bottomrule
\end{tabular}
}
\end{table}

\section{Translations for QA datasets} \label{app:qa_translations} 

We use an in-house translation tool to obtain translations for our datasets. For the question answering tasks (XQuAD and MLQA), the answer span is often not recoverable if the context is translated directly. We experimented with enclosing the answer span in the English context in quotes \cite{Lee2018semi-supervised,Lewis2019mlqa} but found that quotes were often dropped in translations (at different rates depending on the language). We found that enclosing the answer span in HTML tags (e.g. \texttt{<b>} and \texttt{</b>}) worked more reliably. If this fails, as a back-off we fuzzy match the translated answer with the context similar to \cite{Hsu2019zero-shot}. If the minimal edit distance between the closest match and the translated answer is larger than $\min(10, \texttt{answer\_len}/2)$, we drop the example. On the whole, using this combination, we recover more than 97\% of all answer spans in training and test data.

\section{Performance on translated test sets}

We show results comparing the performance of mBERT and translate-train (mBERT) baselines on the XQuAD test sets with automatically translated test sets in Table \ref{tab:xquad-gold-auto-translation-comparison}. Performance on the automatically translated test sets underestimates the performance of mBERT by 2.9 F1 / 0.2 EM points but overestimates the performance of the translate-train baseline by 4.0 F1 / 6.7 EM points. The biggest part of this margin is explained by the difference in scores on the Thai test set. Overall, this indicates that automatically translated test sets are useful as a proxy for cross-lingual performance but may not be reliable for evaluating models that have been trained on translations as these have learnt to exploit the biases of \emph{translationese}.

\begin{table*}[]
\centering
\caption{Comparison of F1 and EM scores of mBERT and translate-train (mBERT) baselines on XQuAD test sets (gold), which were translated by professional translators and automatically translated test sets (auto).}
\label{tab:xquad-gold-auto-translation-comparison}
\resizebox{\textwidth}{!}{%
\begin{tabular}{l c ccccccccccc}
\toprule
 & Test set & es & de & el & ru & tr & ar & vi & th & zh & hi & avg \\ \midrule
\multirow{2}{*}{mBERT} & gold & 75.6 / 56.9 & 70.6 / 54.0 & 62.6 / 44.9 & 71.3 / 53.3 & 55.4 / 40.1 & 61.5 / 45.1 & 69.5 / 49.6 & 42.7 / 33.5 & 58.0 / 48.3 & 59.2 / 46.0 & 62.6 / 47.2 \\
 & auto & 76.1 / 58.7 & 64.3 / 49.9 & 57.9 / 42.5 & 68.3 / 51.8 & 55.6 / 42.9 & 62.1 / 48.6 & 68.6 / 54.3 & 41.1 / 32.6 & 48.5 / 47.7 & 54.1 / 40.9 & 59.7 / 47.0 \\
\multirow{2}{*}{translate-train} & gold & 80.2 / 63.1 & 75.6 / 60.7 & 70.0 / 53.0 & 75.0 / 59.7 & 68.9 / 54.8 & 68.0 / 51.1 & 75.6 / 56.2 & 36.9 / 33.5 & 66.2 / 56.6 & 69.6 / 55.4 & 68.7 / 54.6 \\
 & auto & 80.7 / 66.0 & 71.1 / 58.9 & 69.3 / 54.5 & 75.7 / 61.5 & 71.2 / 59.1 & 74.3 / 60.7 & 76.8 / 64.0 & 79.5 / 74.8 & 59.3 / 58.0 & 69.1 / 55.2 & 72.7 / 61.3 \\
\bottomrule
\end{tabular}%
}
\end{table*}

\begin{table*}[]
\caption{Comparison of accuracy scores of mBERT baseline on XNLI test sets (gold), which were translated by professional translators and automatically translated test sets (auto) in 14 languages. BLEU and chrF scores are computed to measure the translation quality between gold and automatically translated test sets.}
\label{tab:xnli-translated-gold-comparison}
\resizebox{\textwidth}{!}{%
\begin{tabular}{l cccccccccccccc c}
\toprule
Languages & zh    & es    & de    & ar    & ur    & ru    & bg    & el    & fr    & hi    & sw    & th    & tr    & vi    & avg  \\ \midrule
auto Acc. & 69.1  & 74.7  & 72.8  & 66.5  & 64.5  & 71.6  & 70.2  & 67.7  & 74.3  & 65.1  & 50.2  & 54.5  & 60.0  & 72.7  & 66.7       \\ 
gold Acc. & 67.8  & 73.5  & 70.0  & 64.3  & 57.2  & 67.8  & 68.0  & 65.3  & 73.4  & 58.9  & 49.7  & 54.1  & 60.9  & 69.3  & 64.3       \\ \midrule
BLEU      & 40.92 & 43.46 & 30.94 & 32.35 & 20.13 & 22.62 & 45.04 & 60.29 & 47.91 & 29.55 & 31.25 & 10.65 & 15.39 & 56.93 & 34.82      \\ 
chrF      & 35.96 & 67.92 & 60.28 & 59.64 & 48.21 & 50.38 & 67.52 & 75.34 & 69.58 & 53.85 & 59.84 & 54.89 & 51.46 & 69.37 & 58.87      \\ \bottomrule
\end{tabular}}
\end{table*}

\section{mBERT performance across tasks and languages}

\begin{figure}[!t]
    \centering
    \includegraphics[width=1.0\linewidth]{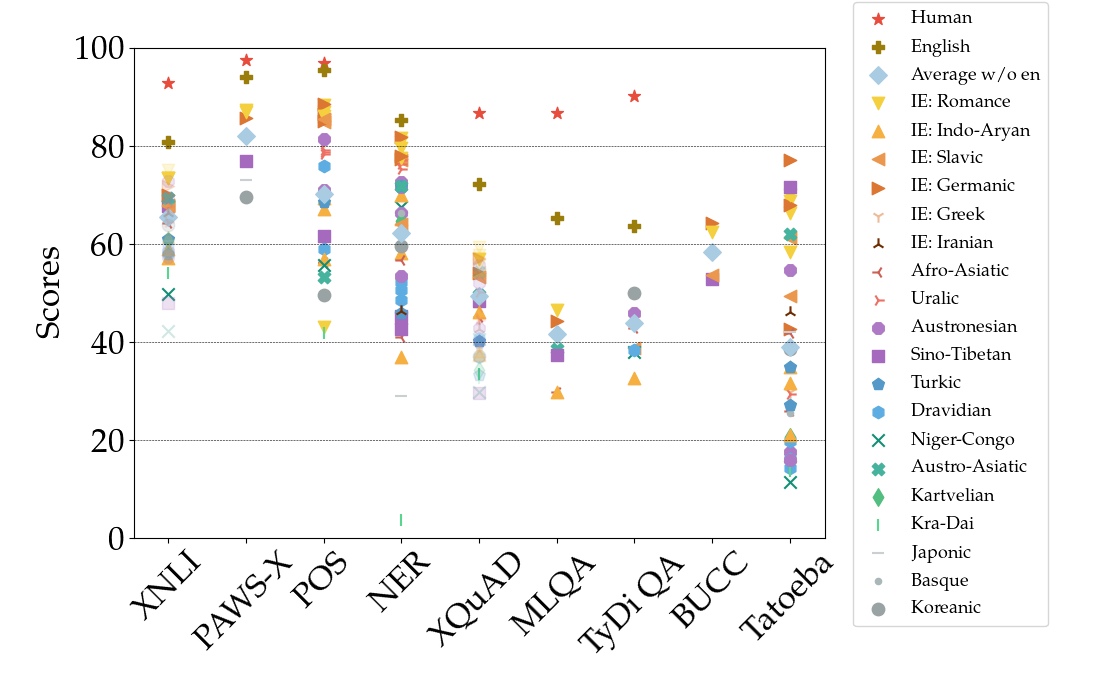}
    \vspace{-1.5mm}
    \caption{An overview of mBERT's performance on the \task tasks for the languages of each task. We highlight an estimate of human performance, performance on the English test set, the average of all languages excluding English, and the family of each language. Performance on pseudo test sets for XNLI and XQuAD is shown with slightly transparent markers.}
    \label{fig:scores_vs_tasks_mbert}
\end{figure}

We show the performance of mBERT across all tasks and languages of \task in Table \ref{fig:scores_vs_tasks_mbert}.

\section{Correlation with pretraining data size}

We show the Pearson correlation coefficient $\rho$ of mBERT, XLM, and XLM-R with the number of Wikipedia articles in Table \ref{tab:dataset-size-correlations}. XLM and mBERT were pretrained on Wikipedia, while XLM-R was pretrained on data from the web.

\begin{table}[]
\centering
\caption{Pearson correlation coefficients ($\rho$) of zero-shot transfer performance and Wikipedia size across datasets and models.}
\label{tab:dataset-size-correlations}
\resizebox{\columnwidth}{!}{%
\begin{tabular}{l c c c c c c c c c}
\toprule
 & XNLI & PAWS-X & POS & NER & XQuAD & MLQA & TyDiQA-GoldP & BUCC & Tatoeba \\ \midrule
mBERT & 0.79 & 0.81 & 0.36 & 0.35 & 0.80 & 0.87 & 0.82 & 0.95 & 0.68 \\
XLM & 0.80 & 0.76 & 0.32 & 0.29 & 0.74 & 0.73 & 0.52 & 0.61 & 0.68 \\
XLM-R & 0.75 & 0.79 & 0.22 & 0.27 & 0.50 & 0.76 & 0.14 & 0.36 & 0.49 \\ \bottomrule
\end{tabular}%
}
\end{table}

\section{Generalization to unseen tag combinations}

\begin{table}[]
\centering
\caption{Accuracy of mBERT on the target language dev data on POS tag trigrams and 4-grams that appeared and did not appear in the English training data. We show the average performance across all non-English languages and the difference of said average compared to the English performance on the bottom.}
\label{tab:pos-tag-generalization}
\begin{tabular}{lllll}
\toprule
 & \begin{tabular}[c]{@{}l@{}}trigram,\\ seen\end{tabular} & \begin{tabular}[c]{@{}l@{}}trigram,\\ unseen\end{tabular} & \begin{tabular}[c]{@{}l@{}}4-gram,\\ seen\end{tabular} & \begin{tabular}[c]{@{}l@{}}4-gram,\\ unseen\end{tabular} \\ \midrule
en & 90.3 & 63.0 & 88.1 & 67.5 \\ \midrule
af & 68.1 & 8.2 & 64.1 & 24.2 \\
ar & 22.0 & 0.7 & 14.9 & 4.6 \\
bg & 63.1 & 14.6 & 56.1 & 23.9 \\
de & 77.8 & 47.2 & 73.0 & 48.7 \\
el & 59.6 & 9.1 & 52.5 & 14.2 \\
es & 68.6 & 10.6 & 62.4 & 24.9 \\
et & 60.7 & 14.4 & 53.1 & 31.9 \\
eu & 32.8 & 7.1 & 28.7 & 8.1 \\
he & 52.7 & 35.7 & 44.0 & 27.4 \\
hi & 38.7 & 13.0 & 32.6 & 12.5 \\
hu & 55.5 & 28.8 & 46.9 & 23.7 \\
id & 60.8 & 16.6 & 54.7 & 21.6 \\
it & 75.5 & 12.8 & 71.8 & 23.5 \\
ja & 16.3 & 0.0 & 12.3 & 1.0 \\
ko & 22.0 & 2.9 & 14.7 & 3.8 \\
mr & 31.7 & 0.0 & 25.5 & 3.3 \\
nl & 75.5 & 24.1 & 71.0 & 37.8 \\
pt & 76.2 & 14.9 & 71.2 & 30.6 \\
ru & 69.1 & 4.8 & 63.8 & 20.6 \\
ta & 30.3 & 0.0 & 24.5 & 4.2 \\
te & 57.8 & 0.0 & 48.7 & 24.7 \\
tr & 41.2 & 6.2 & 33.9 & 10.1 \\
ur & 30.6 & 18.3 & 22.3 & 10.9 \\
zh & 29.0 & 0.0 & 21.7 & 3.9 \\ \midrule
avg & 50.6 & 12.1 & 44.3 & 18.3 \\
diff & 39.7 & 50.9 & 43.7 & 49.2 \\ \bottomrule
\end{tabular}%
\end{table}

We show the performance of mBERT on POS tag trigrams and 4-grams that were seen and not seen in the English training data in Table \ref{tab:pos-tag-generalization}.

\section{Generalization to unseen entities}

\begin{table*}[]
\centering
\caption{Comparison of accuracies for entities in the target language NER dev data that were seen in the English NER training data (a); were not seen in the English NER training data (b); only consist of up to two tokens (c); only consist of Latin characters (d); and occur at least twice in the dev data (e). We only show languages where the sets (a--e) contain at least 100 entities each. We show the difference between (a) and (b) and the minimum difference between (a) and (c-e).}
\label{tab:mbert-entities-ner}
\resizebox{\textwidth}{!}{%
\begin{tabular}{llllllllllllllllllllll}
\toprule
& af & de & el & en & es & et & eu & fi & fr & he & hu & id & it & ka & ms & nl & pt & ru & sw & tr & vi \\ \midrule
(a) Seen & 94.7 & 88.3 & 91.4 & 91.9 & 76.3 & 88.3 & 83.6 & 85.3 & 90.5 & 78.2 & 90.7 & 89.4 & 88.4 & 92.3 & 88.6 & 93.5 & 88.6 & 83.9 & 96.3 & 85.2 & 91.4 \\
(b) Not seen & 82.1 & 80.2 & 74.8 & 84.6 & 80.4 & 78.9 & 69.4 & 79.8 & 80.1 & 56.5 & 78.3 & 58.0 & 81.5 & 70.2 & 75.0 & 82.9 & 82.3 & 68.5 & 66.6 & 73.7 & 73.4 \\ \midrule
(a) $-$ (b) & 12.6 & 8.1 & 16.5 & 7.2 & -4.1 & 9.4 & 14.1 & 5.5 & 10.4 & 21.7 & 12.3 & 31.5 & 6.9 & 22.1 & 13.6 & 10.6 & 6.4 & 15.4 & 29.7 & 11.6 & 18.0 \\ \midrule
(c) Short & 86.5 & 82.9 & 80.3 & 88.2 & 86.6 & 81.7 & 72.5 & 83.9 & 88.6 & 66.3 & 83.7 & 85.8 & 87.2 & 72.5 & 89.1 & 87.6 & 87.8 & 78.0 & 65.7 & 83.1 & 84.6 \\
(d) Latin & 83.6 & 81.2 & 87.5 & 86.2 & 80.0 & 79.5 & 70.3 & 80.3 & 81.1 & 77.2 & 79.9 & 61.8 & 82.6 & 89.6 & 76.3 & 84.2 & 83.0 & 83.8 & 70.0 & 75.0 & 74.9 \\
(e) Freq & 87.3 & 80.6 & 81.9 & 91.6 & 83.4 & 79.4 & 68.8 & 85.7 & 77.3 & 66.8 & 86.0 & 56.5 & 88.8 & 74.3 & 81.3 & 87.1 & 84.4 & 76.5 & 49.1 & 81.9 & 78.6 \\ \bottomrule
$\min($(a) $-$ (c--e)$)$ & 7.4 & 5.4 & 3.9 & 0.3 & 3.7 & 6.6 & 11.0 & 0.4 & 1.9 & 1.0 & 4.7 & 3.6 & 0.4 & 2.7 & 0.5 & 5.9 & 0.8 & 0.1 & 26.4 & 2.2 & 6.8 \\ \bottomrule
\end{tabular}%
}
\end{table*}

\begin{table*}[!h]
\caption{XNLI accuracy scores for each language.}
\label{tab:xnli_results}
\resizebox{\textwidth}{!}{
\begin{tabular}{l|ccccccccccccccc|c}
\toprule
Model                                & en   & ar   & bg   & de   & el   & es   & fr   & hi   & ru   & sw   & th   & tr   & ur   & vi   & zh   & \textbf{avg}  \\
\midrule
mBERT       & 80.8          & 64.3          & 68.0          & 70.0          & 65.3          & 73.5          & 73.4          & 58.9          & 67.8          & 49.7          & 54.1          & 60.9          & 57.2          & 69.3          & 67.8          & 65.4          \\
XLM         & 82.8          & 66.0          & 71.9          & 72.7          & 70.4          & 75.5          & 74.3          & 62.5          & 69.9          & 58.1          & 65.5          & 66.4          & 59.8          & 70.7          & 70.2          & 69.1          \\
XLMR & \textbf{88.7} & \textbf{77.2} & \textbf{83.0} & \textbf{82.5} & \textbf{80.8} & \textbf{83.7} & \textbf{82.2} & \textbf{75.6} & \textbf{79.1} & \textbf{71.2} & \textbf{77.4} & \textbf{78.0} & \textbf{71.7} & \textbf{79.3} & \textbf{78.2} & \textbf{79.2} \\
MMTE        & 79.6          & 64.9          & 70.4          & 68.2          & 67.3          & 71.6          & 69.5          & 63.5          & 66.2          & 61.9          & 66.2          & 63.6          & 60.0          & 69.7          & 69.2          & 67.5         \\
\midrule
\textit{\begin{tabular}[c]{@{}l@{}}Translate-train\\ (multi-task)\end{tabular}}                                                       & 81.9 & \textbf{73.8} & \textbf{77.6} & \textbf{77.6} & \textbf{75.9} & \textbf{79.1} & \textbf{77.8} & 70.7          & \textbf{75.4} & \textbf{70.5} & 70.0            & 74.3          & \textbf{67.4} & \textbf{77.0} & \textbf{77.6} & \textbf{75.1} \\
\textit{Translate-train}  & 80.8          & 73.6          & 76.6          & 77.4          & 75.7          & 78.1          & 77.4          & \textbf{71.9} & 75.2          & 69.4          & \textbf{70.9} & \textbf{75.3} & 67.2          & 75.0        & 74.1          & 74.6          \\
\textit{Translate-test}                                                         & \textbf{85.9}          & 73.1          & 76.6          & 76.9          & 75.3          & 78.0          & 77.5          & 69.1          & 74.8          & 68.0          & 67.1          & 73.5          & 66.4          & 76.6        & 76.3          & 76.8         
\\
\bottomrule
\end{tabular}}
\end{table*}

\begin{table*}[]
\centering
\caption{Tatoeba results (Accuracy) for each language}
\label{tab:tatoeba_results}
\resizebox{\textwidth}{!}{
\begin{tabular}{l|cccccccccccccccccc}
\toprule
Lang. & af            & ar            & bg            & bn            & de            & el            & es            & et            & eu            & fa            & fi            & fr            & he            & hi            & hu            & id            & it            & ja            \\
BERT  & 42.7          & 25.8          & 49.3          & 17            & 77.2          & 29.8          & 68.7          & 29.3          & 25.5          & 46.1          & 39            & 66.3          & 41.9          & 34.8          & 38.7          & 54.6          & 58.4          & 42            \\
XLM   & 43.2          & 18.2          & 40            & 13.5          & 66.2          & 25.6          & 58.4          & 24.8          & 17.1          & 32.2          & 32.2          & 54.5          & 32.1          & 26.5          & 30.1          & 45.9          & 56.5          & 40            \\
XLMR  & \textbf{58.2} & \textbf{47.5} & \textbf{71.6} & \textbf{43}   & \textbf{88.8} & \textbf{61.8} & \textbf{75.7} & \textbf{52.2} & \textbf{35.8} & \textbf{70.5} & \textbf{71.6} & \textbf{73.7} & \textbf{66.4} & \textbf{72.2} & \textbf{65.4} & \textbf{77}   & \textbf{68.3} & \textbf{60.6} \\
\midrule
      & jv            & ka            & kk            & ko            & ml            & mr            & nl            & pt            & ru            & sw            & ta            & te            & th            & tl            & tr            & ur            & vi            & zh            \\
      \midrule
BERT  & 17.6          & 20.5          & 27.1          & 38.5          & 19.8          & 20.9          & 68            & 69.9          & 61.2          & 11.5          & 14.3          & 16.2          & 13.7          & 16            & 34.8          & \textbf{31.6} & 62            & \textbf{71.6} \\
XLM   & \textbf{22.4} & 22.9          & 17.9          & 25.5          & 20.1          & 13.9          & 59.6          & 63.9          & 44.8          & 12.6          & 20.2          & 12.4          & \textbf{31.8} & 14.8          & 26.2          & 18.1          & 47.1          & 42.2          \\
XLMR  & 14.1          & \textbf{52.1} & \textbf{48.5} & \textbf{61.4} & \textbf{65.4} & \textbf{56.8} & \textbf{80.8} & \textbf{82.2} & \textbf{74.1} & \textbf{20.3} & \textbf{26.4} & \textbf{35.9} & 29.4          & \textbf{36.7} & \textbf{65.7} & 24.3          & \textbf{74.7} & 68.3         \\
\bottomrule
\end{tabular}}
\end{table*}

We show the performance of mBERT on entities in the target language NER dev data that were seen and not seen in the English NER training data in Table \ref{tab:mbert-entities-ner}. For simplicity, we count an entity as occurring in the English training data if a subset of at least two tokens matches with an entity in the English training data. As most matching entities in the target language data only consist of up to two tokens, are somewhat frequent, and consist only of Latin characters, we provide the performance on all entities fitting each criterion respectively for comparison. For all target languages in the table except Spanish, entities that appeared in the English training data are more likely to be tagged correctly than ones that did not. The differences are largest for two languages that are typologically distant to English, Indonesian (id) and Swahili (sw). For most languages, entities that appear in the English training data are similarly likely to be correctly classified as entities that are either frequent, appear in Latin characters, or are short. However, for Swahili and Basque (eu), mBERT does much better on entities that appeared in the English training data compared to the comparison entities. Another interesting case is Georgian (ka), which uses a unique script. The NER model is very good at recognizing entities that are written in Latin script but performs less well on entities in Georgian script.

\section{Sentence representations across all layers} 
\begin{figure}
    \centering
    \includegraphics[width=1.0\linewidth]{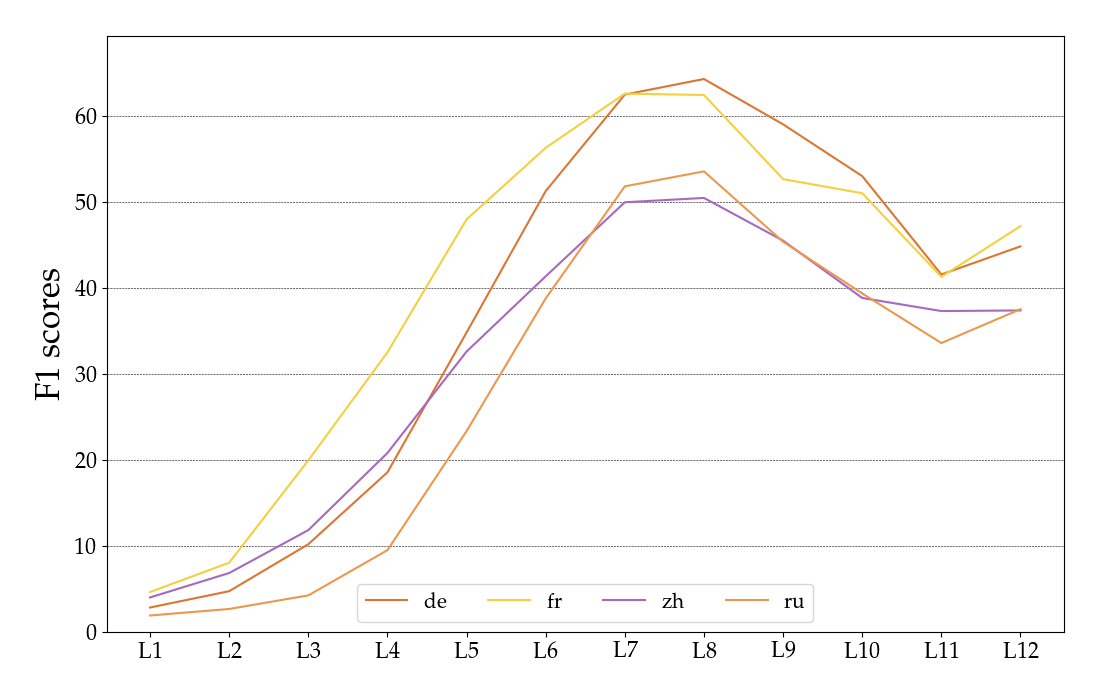}
    \caption{Comparison of mBERT's sentence representations by averaging word embeddings in each layer in the BUCC task.}
    \label{fig:sent_embed}
\end{figure}

$\:$ For sentence retrieval tasks, we analyze whether the multilingual sentence representations obtained from all layers are well-aligned in the embedding spaces. Without fine-tuning on any parallel sentences at all, we explore three ways of extracting the sentence representations from all the models: (1) the embeddings of the first token in the last layer, also known as [CLS] token; (2) the average word embeddings in each layer; (3) the concatenation of the average word embeddings in the bottom, middle, and top 4 layers, i.e., Layer 1 to 4 (bottom), Layer 5 to 8 (middle), Layer 9 to 12 (top). Figure~\ref{fig:sent_embed} shows the F1 scores of the average word embeddings in each layer of mBERT in the BUCC task. We observe that the average word embeddings in the middle layers, e.g., Layer 6 to 8, perform better than that in the bottom or the top layers. In Table~\ref{tab:sent_emb}, we show the performance of these three types of sentence embeddings in the BUCC task. The embeddings of the CLS token perform relatively bad in cross-lingual retrieval tasks. We conjecture that the CLS embeddings highly abstract the semantic meaning of a sentence, while they lose the token-level information which is important for matching two translated sentences in two languages. With respect to the concatenation of average word embeddings from four continuous layers, We also observe that embeddings from the middle layers perform better than that from the bottom and top layers. Average word embeddings in the middle individual layer perform comparative to the concatenated embeddings from the middle four layers.

\begin{table}[]
\caption{Three types of sentence embeddings from mBERT in BUCC tasks: (1) CLS token embeddings in the last layer; (2) Average word embeddings in the middle layers, i.e., Layer 6, 7, 8; (3) the concatenation of average word embeddings in the continuous four layers, i.e., Layer 1-4 (bottom layers), Layer 5-8 (middle layers), Layer 9-12 (top layers).}
\label{tab:sent_emb}
\centering
\begin{tabular}{lrrrr}
\toprule
Type       & de    & fr    & zh    & ru    \\ \midrule
CLS        & 3.88  & 4.73  & 0.89  & 2.15  \\
Layer 6    & 51.29 & 56.32 & 41.38 & 38.81 \\
Layer 7    & 62.51 & 62.62 & 49.99 & 51.84 \\
Layer 8    & 64.32 & 62.46 & 50.49 & 53.58 \\
Layer 1-4  & 6.98  & 12.3  & 12.05 & 4.33 \\
Layer 5-8  & 63.12 & 63.42 & 52.84 & 51.67 \\
Layer 9-12 & 53.97 & 52.68 & 44.18 & 43.13 \\
\bottomrule
\end{tabular}
\end{table}

\subsection{Language Families and Scripts}
\begin{figure*}[!]
\centering
\begin{subfigure}{.5\textwidth}
  \centering
  \includegraphics[width=\linewidth]{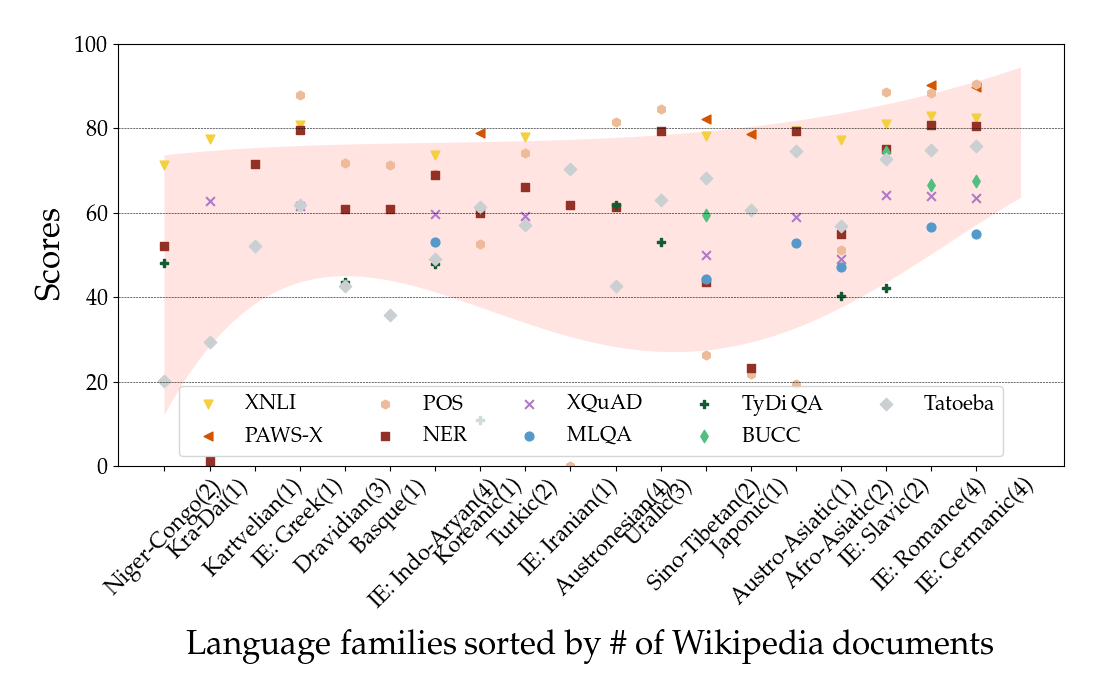}
  \label{fig:xlmr_lang_families}
\end{subfigure}%
\begin{subfigure}{.5\textwidth}
  \centering
  \includegraphics[width=\linewidth]{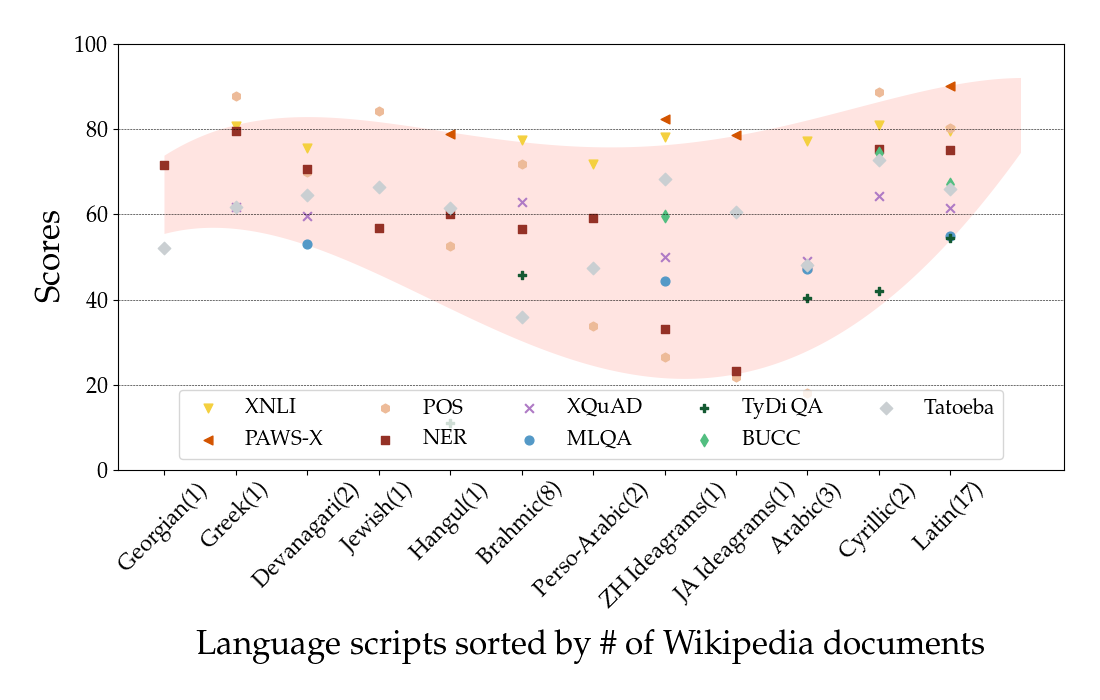}
  \label{fig:xlmr_lang_scripts}
\end{subfigure}
\vspace{-0.8cm}
\caption{Performance of XLM-R across tasks grouped by language families (left) and scripts (right). The number of languages per group is in brackets and the groups are from low-resource to high-resource on the x-axis. We additionally plot the 3rd order polynomial fit for the minimum and maximum values for each group.}
\label{fig:xlmr_correlation_lang_family_script}
\end{figure*}

We also report the performance of XLM-R in all the tasks across different language families and writing scripts in Figure~\ref{fig:xlmr_correlation_lang_family_script}.
\begin{table}[!ht]
\caption{PAWS-X accuracy scores for each language.}
\label{tab:paws-x-results}
\resizebox{\columnwidth}{!}{
\begin{tabular}{l|ccccccc|c}
\toprule
Model                                                                           & en   & de   & es   & fr   & ja   & ko   & zh   & \textbf{avg}  \\
\midrule
mBERT       & 94.0          & 85.7          & 87.4          & 87.0          & 73.0          & 69.6          & 77.0          & 81.9          \\
XLM         & 94.0          & 85.9          & 88.3          & 87.4          & 69.3          & 64.8          & 76.5          & 80.9          \\
XLMR & \textbf{94.7} & \textbf{89.7} & \textbf{90.1} & \textbf{90.4} & \textbf{78.7} & \textbf{79.0} & \textbf{82.3} & \textbf{86.4} \\
MMTE        & 93.1          & 85.1          & 87.2          & 86.9          & 72.0          & 69.2          & 75.9          & 81.3          \\
\midrule
\textit{Translate-train}                                                        & 94.0          & 87.5          & 89.4          & 89.6          & 78.6          & 81.6          & 83.5          & 86.3          \\
\textit{\begin{tabular}[c]{@{}l@{}}Translate-train\\ (multi-task)\end{tabular}} & \textbf{94.5} & \textbf{90.5} & \textbf{91.6} & \textbf{91.7} & \textbf{84.4} & \textbf{83.9} & \textbf{85.8} & \textbf{88.9} \\
\textit{Translate-test}                                                         & 93.5          & 88.2          & 89.3          & 87.4          & 78.4          & 76.6          & 77.6          & 84.4         \\
\bottomrule
\end{tabular}}
\end{table}

\begin{table}[h]
\centering
\caption{BUCC results (F1 scores) for each languages.}
\label{tab:bucc_results}
\begin{tabular}{l|cccc|c}
\toprule
Model & de            & fr            & ru   & zh            & avg           \\ 
\midrule
BERT  & 62.5          & 62.6          & 51.8 & 50.0          & 56.7          \\
XLM   & 56.3          & 63.9          & 60.6 & 46.6          & 56.8          \\
XLMR  & 67.5          & \textbf{66.5} & \textbf{73.5} & \textbf{56.7} & \textbf{66.0} \\
MMTE  & \textbf{67.9} & 63.9          & 54.3 & 53.3          & 59.8          \\
\bottomrule
\end{tabular}
\end{table}

\section{Results for each task and language}

We show the detailed results for all tasks and languages in Tables \ref{tab:xnli_results} (XNLI), \ref{tab:paws-x-results} (PAWS-X), \ref{tab:pos_results} (POS), \ref{tab:ner_results} (NER), \ref{tab:xquad-results} (XQuAD), \ref{tab:mlqa_results} (MLQA), \ref{tab:tydiqa_results} (TyDiQA-GoldP), \ref{tab:bucc_results} (BUCC), and \ref{tab:tatoeba_results} (Tatoeba).

\begin{table*}[]
\caption{XQuAD results (F1 / EM) for each language.}
\label{tab:xquad-results}
\resizebox{\textwidth}{!}{
\begin{tabular}{l|ccccccccccc|c}
\toprule
Model                                                                           & en          & ar          & de          & el          & es          & hi          & ru          & th          & tr          & vi          & zh          & \textbf{avg}         \\
\midrule
mBERT       & 83.5 / 72.2          & 61.5 / 45.1          & 70.6 / 54.0          & 62.6 / 44.9          & 75.5 / 56.9          & 59.2 / 46.0          & 71.3 / 53.3          & 42.7 / 33.5          & 55.4 / 40.1          & 69.5 / 49.6          & 58.0 / 48.3          & 64.5 / 49.4          \\
XLM         & 74.2 / 62.1          & 61.4 / 44.7          & 66.0 / 49.7          & 57.5 / 39.1          & 68.2 / 49.8          & 56.6 / 40.3          & 65.3 / 48.2          & 35.4 / 24.5          & 57.9 / 41.2          & 65.8 / 47.6          & 49.7 / 39.7          & 59.8 / 44.3          \\
XLMR & \textbf{86.5 / 75.7} & \textbf{68.6 / 49.0} & \textbf{80.4 / 63.4} & \textbf{79.8 / 61.7} & \textbf{82.0 / 63.9} & \textbf{76.7 / 59.7} & \textbf{80.1 / 64.3} & \textbf{74.2 / 62.8} & \textbf{75.9 / 59.3} & \textbf{79.1 / 59.0} & \textbf{59.3 / 50.0} & \textbf{76.6 / 60.8} \\
MMTE        & 80.1 / 68.1          & 63.2 / 46.2          & 68.8 / 50.3          & 61.3 / 35.9          & 72.4 / 52.5          & 61.3 / 47.2  & 68.4 / 45.2          & 48.4 / 35.9          & 58.1 / 40.9          & 70.9 / 50.1          & 55.8 / 36.4          & 64.4 / 46.2          \\
\midrule
\textit{Translate-train}                                                        & 83.5 / 72.2          & 68.0 / 51.1          & 75.6 / 60.7          & 70.0 / 53.0          & 80.2 / 63.1          & 69.6 / 55.4          & 75.0 / 59.7          & 36.9 / 33.5          & 68.9 / 54.8          & 75.6 / 56.2          & 66.2 / 56.6          & 70.0 / 56.0          \\
\textit{\begin{tabular}[c]{@{}l@{}}Translate-train\\ (multi-task)\end{tabular}} & 86.0 / 74.5          & 71.0 / 54.1          & 78.8 / 63.9          & 74.2 / 56.1          & 82.4 / 66.2          & 71.3 / 56.2          & 78.1 / 63.0          & 38.1 / 34.5          & 70.6 / 55.7          & 78.5 / 58.8          & 67.7 / 58.7          & 72.4 / 58.3          \\
\textit{Translate-test}                                                         & \textbf{87.9 / 77.1} & \textbf{73.7 / 58.8} & \textbf{79.8 / 66.7} & \textbf{79.4 / 65.5} & \textbf{82.0 / 68.4} & \textbf{74.9 / 60.1} & \textbf{79.9 / 66.7} & \textbf{64.6 / 50.0} & \textbf{67.4 / 49.6} & \textbf{76.3 / 61.5} & \textbf{73.7 / 59.1} & \textbf{76.3 / 62.1} \\
\bottomrule
\end{tabular}}

\end{table*}

\begin{table*}[]
\centering
\caption{TyDiQA-GoldP results (F1 / EM) for each language.}
\label{tab:tydiqa_results}
\resizebox{\textwidth}{!}{%
\begin{tabular}{l c c c c c c c c c c}
\toprule
Model & en & ar & bn & fi & id & ko & ru & sw & te & avg \\ \midrule
mBERT & \textbf{75.3 / 63.6} & 62.2 / 42.8 & 49.3 / 32.7 & 59.7 / 45.3 & 64.8 / 45.8 & \textbf{58.8 / 50.0} & 60.0 / 38.8 & 57.5 / 37.9 & 49.6 / 38.4 & 59.7 / 43.9 \\
XLM & 66.9 / 53.9 & 59.4 / 41.2 & 27.2 / 15.0 & 58.2 / 41.4 & 62.5 / 45.8 & 14.2 / 5.1 & 49.2 / 30.7 & 39.4 / 21.6 & 15.5 / 6.9 & 43.6 / 29.1 \\
XLM-R & 71.5 / 56.8 & \textbf{67.6 / 40.4} & \textbf{64.0 / 47.8} & \textbf{70.5 / 53.2} & \textbf{77.4 / 61.9} & 31.9 / 10.9 & \textbf{67.0 / 42.1} & \textbf{66.1 / 48.1} & \textbf{70.1 / 43.6} & \textbf{65.1 / 45.0} \\
MMTE & 62.9 / 49.8 & 63.1 / 39.2 & 55.8 / 41.9 & 53.9 / 42.1 & 60.9 / 47.6 & 49.9 / 42.6 & 58.9 / 37.9 & 63.1 / 47.2 & 54.2 / 45.8 & 58.1 / 43.8 \\ \midrule
\textit{Translate-train} & 75.3 / 63.6 & 61.5 / 44.1 & 31.9 / 31.9 & 62.6 / 49.0 & 68.6 / 52.0 & 53.2 / 41.3 & 53.1 / 33.9 & 61.9 / 45.5 & 27.4 / 17.5 & 55.1 / 42.1 \\
\textit{\begin{tabular}[c]{@{}l@{}}Translate-train\\ (multi-task)\end{tabular}} & 73.2 / 62.5 & \textbf{71.8 / 54.2} & 49.7 / 36.3 & 68.1 / 53.6 & 72.3 / 55.2 & 58.6 / 47.8 & 64.3 / 45.3 & 66.8 / 48.9 & 53.3 / 40.2 & 64.2 / 49.3 \\
\textit{Translate-test} & \textbf{75.9 / 65.9} & 68.8 / 49.6 & \textbf{66.7 / 48.1} & \textbf{72.0 / 56.6} & \textbf{76.8 / 60.9} & \textbf{69.2 / 55.7} & \textbf{71.4 / 54.3} & \textbf{73.3 / 53.8} & \textbf{75.1 / 59.2} & \textbf{72.1 / 56.0} \\ \midrule
\textit{Monolingual} & 75.3 / 63.6 & 80.5 / 67.0 & 71.1 / 60.2 & 75.6 / 64.1 & \textbf{81.3 / 70.4} & 59.0 / 49.6 & 72.1 / 56.2 & 75.0 / 66.7 & 80.2 / 66.4 & 74.5 / 62.7 \\
\textit{\begin{tabular}[c]{@{}l@{}}Monolingual\\ few-shot\end{tabular}} & 63.1 / 50.9 & 61.3 / 44.8 & 58.7 / 49.6 & 51.4 / 38.1 & 70.4 / 58.1 & 45.4 / 38.4 & 56.9 / 42.6 & 55.4 / 46.3 & 65.2 / 49.6 & 58.7 / 46.5 \\
\textit{\begin{tabular}[c]{@{}l@{}}Joint\\ monolingual\end{tabular}} & \textbf{77.6 / 69.3} & \textbf{82.7 / 69.4} & \textbf{79.6 / 69.9} & \textbf{79.2 / 67.8} & 68.9 / 72.7 & \textbf{68.9 / 59.4} & \textbf{75.8 / 59.2} & \textbf{81.9 / 74.3} & \textbf{83.4 / 70.3} & \textbf{77.6 / 68.0} \\
\bottomrule
\end{tabular}%
}
\end{table*}

\begin{table*}[]
\centering
\caption{MLQA results (F1 / EM) for each language.}
\label{tab:mlqa_results}
\resizebox{0.8\textwidth}{!}{
\begin{tabular}{l|ccccccc|c}
\toprule
Model  & en & ar & de & es & hi & vi & zh                   & avg                  \\
\toprule
mBERT   & 80.2 / 67.0          & 52.3 / 34.6          & 59.0 / 43.8          & 67.4 / 49.2          & 50.2 / 35.3          & 61.2 / 40.7          & 59.6 / 38.6          & 61.4 / 44.2          \\
XLM & 68.6 / 55.2 & 42.5 / 25.2 & 50.8 / 37.2 & 54.7 / 37.9  & 34.4 / 21.1  & 48.3 / 30.2 & 40.5 / 21.9 & 48.5 / 32.6 \\
XLM-R                                                                     & \textbf{83.5 / 70.6} & \textbf{66.6 / 47.1} & \textbf{70.1 / 54.9} & \textbf{74.1 / 56.6} & \textbf{70.6 / 53.1} & \textbf{74 / 52.9}   & \textbf{62.1 / 37.0} & \textbf{71.6 / 53.2} \\
MMTE                                                                            & 78.5 / --            & 56.1 / --            & 58.4 / --            & 64.9 / --            & 46.2 / --            & 59.4 / --            & 58.3 / --            & 60.3 / 41.4          \\
\midrule
\textit{Translate-train}                                                        & 80.2 / 67.0          & 55.0 / 35.6          & 64.4 / 49.4          & 70.0 / 52.0          & 60.1 / 43.4          & 65.7 / 45.5          & 63.9 / 42.7          & 65.6 / 47.9          \\
\textit{\begin{tabular}[c]{@{}l@{}}Translate-train\\ (multi-task)\end{tabular}} & 80.7 / 67.7          & 58.9 / 39.0          & 66.0 / 51.6          & 71.3 / 53.7          & 62.4 / 45.0          & 67.9 / 47.6          & 66.0 / 43.9          & 67.6 / 49.8          \\
\textit{Translate-test}                                                         & \textbf{83.8 / 71.0} & \textbf{65.3 / 46.4} & \textbf{71.2 / 54.0} & \textbf{73.9 / 55.9} & \textbf{71.0 / 55.1} & \textbf{70.6 / 54.0} & \textbf{67.2 / 50.6} & \textbf{71.9 / 55.3} \\
\bottomrule
\end{tabular}}
\end{table*}

\begin{table*}[]
\centering
\caption{POS results (Accuracy) for each language}
\label{tab:pos_results}
\resizebox{\textwidth}{!}{
\begin{tabular}{l|ccccccccccccccccc}
\toprule
Lang. & af            & ar            & bg            & de            & el            & en            & es            & et            & eu            & fa            & fi            & fr            & he            & hi            & hu            & id            & it            \\
\midrule
mBERT & 86.6          & 56.2          & 85.0          & 85.2          & 81.1          & 95.5          & 86.9          & 79.1          & 60.7          & 66.7          & 78.9          & 84.2          & 56.2          & 67.2          & 78.3          & 71.0          & 88.4          \\
XLM   & 88.5          & 63.1          & 85.0          & 85.8          & 84.3          & 95.4          & 85.8          & 78.3          & 62.8          & 64.7          & 78.4          & 82.8          & 65.9          & 66.2          & 77.3          & 70.2          & 87.4          \\
XLMR  & \textbf{89.8} & \textbf{67.5} & \textbf{88.1} & \textbf{88.5} & \textbf{86.3} & 96.1          & \textbf{88.3} & \textbf{86.5} & \textbf{72.5} & \textbf{70.6} & \textbf{85.8} & \textbf{87.2}          & \textbf{68.3} & \textbf{76.4} & \textbf{82.6} & 72.4          & \textbf{89.4} \\
MMTE  & 86.2          & 65.9          & 87.2          & 85.8          & 77.7          & \textbf{96.6} & 85.8          & 81.6          & 61.9          & 67.3          & 81.1          & 84.3 & 57.3          & 76.4          & 78.1          & \textbf{73.5} & 89.2          \\
\midrule
   & ja            & kk            & ko            & mr            & nl            & pt            & ru            & ta            & te            & th            & tl            & tr            & ur            & vi            & yo            & zh            & avg           \\
      \midrule
mBERT & \textbf{49.2} & 70.5          & 49.6          & 69.4          & 88.6          & 86.2          & 85.5          & 59.0          & 75.9          & 41.7          & 81.4          & 68.5          & 57.0          & 53.2          & \textbf{55.7} & 61.6          & 71.5          \\
XLM   & 49.0          & 70.2          & 50.1          & 68.7          & 88.1          & 84.9          & 86.5          & 59.8          & 76.8          & 55.2          & 76.3          & 66.4          & 61.2          & 52.4          & 20.5          & 65.4          & 71.3          \\
XLMR  & 15.9          & \textbf{78.1} & 53.9          & \textbf{80.8} & \textbf{89.5} & \textbf{87.6} & \textbf{89.5} & \textbf{65.2} & \textbf{86.6} & \textbf{47.2} & \textbf{92.2} & \textbf{76.3} & \textbf{70.3} & \textbf{56.8} & 24.6          & 25.7          & \textbf{73.8} \\
MMTE  & 48.6          & 70.5          & \textbf{59.3} & 74.4          & 83.2          & 86.1          & 88.1          & 63.7          & 81.9          & 43.1          & 80.3          & 71.8          & 61.1          & 56.2          & 51.9          & \textbf{68.1} & 73.5  \\
\bottomrule
\end{tabular}}
\end{table*}

\begin{table*}[]
\centering
\caption{NER results (F1 Score) for each language}
\label{tab:ner_results}
\resizebox{\textwidth}{!}{
\begin{tabular}{l|cccccccccccccccccccc}
\toprule
Lang. & en            & af            & ar            & bg            & bn            & de            & el            & es            & et            & eu            & fa            & fi            & fr            & he            & hi            & hu            & id            & it            & ja            & jv            \\
\midrule
mBERT & \textbf{85.2} & 77.4          & 41.1          & 77.0          & 70.0          & 78.0          & 72.5          & 77.4          & 75.4          & \textbf{66.3} & 46.2          & 77.2          & 79.6          & 56.6          & 65.0          & 76.4          & \textbf{53.5} & \textbf{81.5} & 29.0          & \textbf{66.4} \\
XLM   & 82.6          & 74.9          & 44.8          & 76.7          & 70.0          & 78.1          & 73.5          & 74.8          & 74.8          & 62.3          & 49.2          & 79.6          & 78.5          & 57.7          & 66.1          & 76.5          & 53.1          & 80.7          & 23.6          & 63.0          \\
XLMR  & 84.7          & \textbf{78.9} & \textbf{53.0} & \textbf{81.4} & \textbf{78.8} & \textbf{78.8} & \textbf{79.5} & \textbf{79.6} & \textbf{79.1} & 60.9          & \textbf{61.9} & \textbf{79.2} & \textbf{80.5} & \textbf{56.8} & \textbf{73.0} & \textbf{79.8} & 53.0          & 81.3          & 23.2          & 62.5          \\
MMTE  & 77.9          & 74.9          & 41.8          & 75.1          & 64.9          & 71.9          & 68.3          & 71.8          & 74.9          & 62.6          & 45.6          & 75.2          & 73.9          & 54.2          & 66.2          & 73.8          & 47.9          & 74.1          & \textbf{31.2} & 63.9          \\
\midrule
    & ka            & kk            & ko            & ml            & mr            & ms            & my            & nl            & pt            & ru            & sw            & ta            & te            & th            & tl            & tr            & ur            & vi            & yo            & zh            \\
      \midrule
mBERT & 64.6          & 45.8          & 59.6          & 52.3          & 58.2          & \textbf{72.7} & 45.2          & 81.8          & 80.8          & 64.0          & 67.5          & 50.7          & 48.5          & 3.6           & 71.7          & 71.8          & 36.9          & 71.8          & \textbf{44.9} & \textbf{42.7} \\
XLM   & 67.7          & \textbf{57.2} & 26.3          & 59.4          & 62.4          & 69.6          & 47.6          & 81.2          & 77.9          & 63.5          & 68.4          & 53.6          & 49.6          & 0.3           & \textbf{78.6} & 71.0          & 43.0          & 70.1          & 26.5          & 32.4          \\
XLMR  & \textbf{71.6} & 56.2          & \textbf{60.0} & \textbf{67.8} & \textbf{68.1} & 57.1          & \textbf{54.3} & \textbf{84.0} & \textbf{81.9} & \textbf{69.1} & \textbf{70.5} & \textbf{59.5} & \textbf{55.8} & 1.3           & 73.2          & \textbf{76.1} & \textbf{56.4} & \textbf{79.4} & 33.6          & 33.1          \\
MMTE  & 60.9          & 43.9          & 58.2          & 44.8          & 58.5          & 68.3          & 42.9          & 74.8          & 72.9          & 58.2          & 66.3          & 48.1          & 46.9          & \textbf{3.9}  & 64.1          & 61.9          & 37.2          & 68.1          & 32.1          & 28.9         \\
\bottomrule
\end{tabular}}
\end{table*}

\end{document}